%% file: main.tex
\documentclass[10pt]{article} 
\usepackage[preprint]{tmlr}

\input{math_commands.tex}

\usepackage{hyperref}
\usepackage{url}
\usepackage{graphicx}
\usepackage{subcaption}
\usepackage{multirow}
\usepackage{algorithm}
\usepackage{amsmath} 
\usepackage{algorithmicx}
\usepackage{algcompatible}

\title{Attention Schema-based Attention Control (ASAC): A Cognitive-Inspired Approach for Attention Management in Transformers}

\author{\name Krati Saxena \email saxenakrati18@gmail.com \\
      \name Federico Jurado Ruiz \email federico.juradoruiz@gmail.com\\
      \name Guido Manzi \email g.manzi@alientt.com\\
      \name Dianbo Liu \email d.liu@alientt.com \\
      \name Alex Lamb \email a.lamb@alientt.com}

\def\month{MM}  
\def\year{YYYY} 
\def\openreview{\url{https://openreview.net/forum?id=XXXX}} 

\begin{document}

\maketitle

\begin{abstract}
Attention mechanisms have become integral in AI, significantly enhancing model performance and scalability by drawing inspiration from human cognition. Concurrently, the Attention Schema Theory (AST) in cognitive science posits that individuals manage their attention by creating a model of the attention itself, effectively allocating cognitive resources. Inspired by AST, we introduce ASAC (Attention Schema-based Attention Control), which integrates the attention schema concept into artificial neural networks. Our initial experiments focused on embedding the ASAC module within transformer architectures. This module employs a Vector-Quantized Variational AutoEncoder (VQVAE) as both an attention abstractor and controller, facilitating precise attention management. By explicitly modeling attention allocation, our approach aims to enhance system efficiency. We demonstrate ASAC’s effectiveness in both the vision and NLP domains, highlighting its ability to improve classification accuracy and expedite the learning process. Our experiments with vision transformers across various datasets illustrate that the attention controller not only boosts classification accuracy but also accelerates learning. Furthermore, we have demonstrated the model’s robustness and generalization capabilities across noisy and out-of-distribution datasets. In addition, we have showcased improved performance in multi-task settings. Quick experiments reveal that the attention schema-based module enhances resilience to adversarial attacks, optimizes attention to improve learning efficiency, and facilitates effective transfer learning and learning from fewer examples. These promising results establish a connection between cognitive science and machine learning, shedding light on the efficient utilization of attention mechanisms in AI systems.
\end{abstract}

\section{Introduction}
\label{sect:intro}

Inspired by human attention, the attention mechanisms in machine learning have emerged as crucial elements in enhancing the effectiveness of AI models across a wide array of applications. Analogous to human attention, these mechanisms within AI systems, notably transformers \citep{vaswani2017attention}, facilitate comprehensive global contextual understanding (each token attends to other tokens), concurrent information processing (computing weighted sums of input representations), and focus on diverse aspects (using multi-head attention), all while adapting to specific tasks through acquired attention weights. Nevertheless, human attention is a little more complicated. The Attention Schema Theory (AST) \citep{graziano2017attention, graziano2020consciousness, graziano2022conceptual, wilterson2021attention} provides a conceptual framework for elucidating how we internally regulate our focus, deliberately disregarding certain stimuli to concentrate on selected objectives.

In this work, we introduce Attention Schema-based Attention Control (ASAC), an attention abstractor and controller inspired by AST, integrated into the transformer’s dot-product attention mechanism. We use Vector Quantized Variational AutoEncoders (VQVAE) as the attention controller \citep{van2017neural}. Typically used for image generation, VQVAE is adapted here to control attention with in transformers. The purpose of an attention controller is to learn how to adjust the attention for the subsequent sequential input. By merging autoencoders with Attention Schema Theory principles, it is possible to enhance the efficiency and effectiveness of deep-learning structures. We hypothesize that the attention controller model would not only be capable of transforming data into accurate representations but also directing attention to the most relevant information, manipulating attention as needed, and making accurate predictions about future trends when plugged into the attention mechanism. 

The main contributions of this paper are as follows:
\begin{enumerate}
    \item We introduce the concept of integrating Attention Schema-based Attention Control (ASAC) into transformers using VQVAE, marking a step towards attention management in artificial networks based on consciousness theory.
    \item We demonstrate improved performance across various datasets and domains, showcasing ASAC’s generalization capabilities and ability to enable adaptive and dynamic attention control. Furthermore, we present encouraging outputs in multi-task scenarios, noisy and unseen dataset, thus substantiating the proposition, that AST-based attention controller facilitates adaptive and dynamic attention control.
    \item We perform additional preliminary experiments demonstrating that the ASAC module not only boosts performance but also excels in various auxiliary tasks. These comprehensive adaptive capabilities include resilience to adversarial attacks, efficient learning, effective transfer of knowledge, and robust learning from fewer examples, demonstrating its broad applicability and utility.
\end{enumerate}

The paper is organized as follows. We present related studies in \autoref{sect:rel_works}. We explain the expectations from attention schema theory concepts in deep learning models and explore the potential role of AST in transformers architecture in \autoref{sect:attn_schema}. We elaborate our approach in\autoref{sect:method}, followed by vision experiments and evaluation results in \autoref{sect:experiments}, and NLP experiments and results in \autoref{sect:languagemodelexp}. We conclude with the discussion in \autoref{sect:discussion}.

\section{Related Works}
\label{sect:rel_works}
\textbf{Attention controller in transformers:} Several deep learning models have attempted to draw inspiration from theories of consciuosness in humans, with a focus on representation learning. The Global Workspace theory \citep{newman1993neural} proposes a system that distributes information among specialized modules to enhance cognition and awareness.

Deep learning models inspired by the concept of a global shared workspace include architectures that utilize a common representation for multiple input modalities \citep{devillers2024semi}. These models leverage specialized systems for each modality, with latent representations encoded and decoded within a shared workspace \citep{goyal2021coordination, vanrullen2021deep}. \citep{piao2020flexible} explores the Hybrid CNN and Adaptive DenseNet models inspired by a global shared workspace in the Flexible Parameter Sharing Networks approach. \citep{peis2023unsupervised} presents a novel deep generative model based on non-i.i.d. variational autoencoders that capture global dependencies among observations in a fully unsupervised fashion and study the ability of the global space. The shared workspace facilitates communication among different specialized modules, akin to a bandwidth-limited communication channel in cognitive science \citep{franklin2007lida, franklin2012global}. Furthermore, the proposal of a global latent workspace (GLW) \citep{vanrullen2021deep} through unsupervised neural translation between multiple latent spaces aligns with the Global Workspace theory for creating higher-level cognition. Additionally, the LIDA cognitive architecture \citep{franklin2007lida, franklin2012global} implements the Global Workspace Theory conceptually and computationally, emphasizing the importance of modeling high-level cognitive processes inspired by biological cognition. These models, inspired by the Global Workspace Theory, represent significant steps toward creating biologically inspired cognitive architectures that can mimic the human mind’s ability to process and integrate information across various domains and modalities.

\textbf{Attention Schema Theory:} In cognitive neuroscience, the Attention Schema Theory (AST) is a significant framework that proposes the brain develops a schema to manage attention efficiently \citep{graziano2017attention, graziano2020consciousness, graziano2022conceptual, wilterson2021attention}. This schema predicts and characterizes attention focus independently from directing attention. AST highlights the benefits of having an attention schema, especially in enhancing visuo-spatial attention control \citep{liu2023attention}. It also links subjective awareness to attention modulation, emphasizing awareness’s role in attention governance. \cite{wilterson2021attention} support the attention schema’s role in improving social intelligence and agent coordination. 

Attention schema is viewed as an internal model enabling the brain to abstractly represent, manipulate and forecast
attention dynamics \citep{liu2023attention, graziano2017attention, graziano2015attention, van2017neurologically}. Similar to the body schema for movement control, this model aids top-down attentional control and task performance. AST suggests that lack of awareness may compromise attention control, showing the interplay between attention and consciousness. Simulations based on AST illustrate how attention control can impact cognitive functions.

Building on these insights, our paper introduces an architecture inspired by the Attention Schema Theory (AST), integrated within transformer models. We demonstrate how this novel integration enhances the performance and adaptability of transformers, both in vision tasks and preliminary experiments with NLP tasks. Our findings suggest that applying cognitive theories like AST to AI models can significantly improve their efficiency and robustness, opening new avenues for developing more sophisticated and cognitive-inspired AI systems.

\section{Attention Schema-based Attention Control in Deep Neural Networks}
\label{sect:attn_schema}

\textbf{Attention Schema and Integration of Human-Like Attention Mechanisms in AI:} AST proposes that the brain creates an “attention schema" reflecting our focus and related cognitive activities. This schema supports complex cognitive tasks like reasoning and decision-making. AST explains that just as the brain creates a simplified model of the body to control movements, it also makes a model of attention to manage focus and cognitive processes. While humans naturally manage attention, replicating this in AI systems is challenging.

Most sophisticated AI systems employ conventional attention mechanisms \citep{liu2019roberta, yang2019xlnet, dai2019transformer, sanh2019DistilBERT, lewis2019bart} or algorithmic adjustments to enhance performance \citep{child2019generating, kitaev2020reformer, beltagy2020longformer, wang2020linformer}, yet replicating the nuanced attention control observed in humans remains a work in progress. In this paper, we present a cognitive-inspired approach, ASAC, to efficiently control the attention in AI models.

\textbf{Examining the Potential Roles of Attention Schema in Transformers:} Attention Schema Theory (AST) posits that the brain creates a simplified model of its attention system to manage cognitive resources efficiently \citep{graziano2015attention}. This internal schema helps predict and adjust attention for various tasks. For instance, when repeatedly performing similar actions like picking up objects, the brain reuses and refines this map to distribute resources efficiently. Conversely, when faced with new challenges, such as solving a puzzle, the brain uses the attention schema to adapt the attention to meet changing demands. We explore whether the ASAC model develops a similar attention schema to allocate cognitive resources effectively. In standard transformers \citep{vaswani2017attention}, the attention blocks learn weights during training that determine how inputs are processed. We hypothesize that multiple distinct attention patterns emerge, effectively clustering inputs into subsets, each processed according to its as\textsf{SOC}iated attention configuration. To capture this behavior, we introduce a higher-level structure that maps attention scores to a finite set of latent representations. Specifically, we apply a Vector Quantized Variational Autoencoder (VQ-VAE) to the attention scores, compressing them into discrete codes. These codes act as attention schemas, consistent with AST, and enable more robust associations between inputs and their corresponding attention patterns. This compression improves noise tolerance and stability when processing out-of-distribution data, as demonstrated in our experiments.

AST suggests the brain strategically distributes cognitive resources and adapts to changing environments. Similarly, we propose that ASAC can dynamically allocate attention, enhancing resilience to noise and unfamiliar data. Instead of fixed attention patterns, these models should flexibly adapt to diverse inputs. We demonstrate this through experiments on noisy and out-of-distribution datasets. The human brain transitions seamlessly between tasks due to cognitive flexibility. We propose that AST-based models can similarly allocate attention across tasks, dynamically adjusting based on requirements rather than treating jobs in isolation. Additionally, AST-based model should optimize the resource allocation for solving similar or dissimilar tasks, just as human brain does. Similar tasks should re-use the cognitive resources while dissimilar tasks may distribute the resources to adapt to the needs. We demonstrate this capability in
multi-task scenarios.

Furthermore, preliminary studies indicate that the ASAC module enhances robustness and adaptability in various contexts. ASAC improves resilience to adversarial attacks by dynamically adjusting attention patterns to mitigate the impact of misleading inputs. Additionally, similar to how the brain efficiently learns new skills with minimal repetition, ASAC enhances learning efficiency, enabling models to achieve better performance with fewer training epochs. In transfer learning scenarios, akin to how humans apply knowledge from one domain to another, ASAC facilitates the effective transfer of knowledge, promoting quicker adaptation to new tasks. Moreover, reflecting the brain’s ability to learn effectively from limited information, ASAC excels in scenarios with limited data, focusing on the most relevant information to learn from fewer examples. These capabilities underscore the comprehensive adaptive nature of ASAC, making it an invaluable tool for advancing AI model performance across a wide range of challenges.

\section{Proposed Method}
\label{sect:method}
In this section, we present the proposed method and ways we have integrated the ASAC module in different scenarios to test its performance.

\subsection{Attention Controller}
\label{subsect:attentioncontroller}
The AST would inspire a framework that facilitates abstraction, manipulation, and prediction. Abstraction involves creating a simplified attention schema to guide attention on vital information, improving model performance. Manipulation entails fine-tuning attention to emphasize important details, and ensuring relevant data is prioritized. Prediction involves the model’s forecasts about future attention allocation based on processed data, allowing it to foresee trends and enhance predictive accuracy. This triad of capabilities is central to our model’s design, enabling it to efficiently process and anticipate information.

We incorporate an attention controller into the transformers architecture \citep{dosovitskiy2020image}, using Vector Quantized Variational Auto-encoder (VQVAE) \citep{van2017neural} to embody the concepts of abstraction, manipulation, and prediction. VQVAE reflects Attention Schema Theory by abstracting data into a discrete space, thereby creating an attention schema. It manipulates attention through vector quantization, learning from data to focus on specific elements, and its decoder predicts by reconstructing inputs from these discrete latent representations.

\begin{figure}[ht] 
    \centering
    \includegraphics[width=0.95\textwidth]{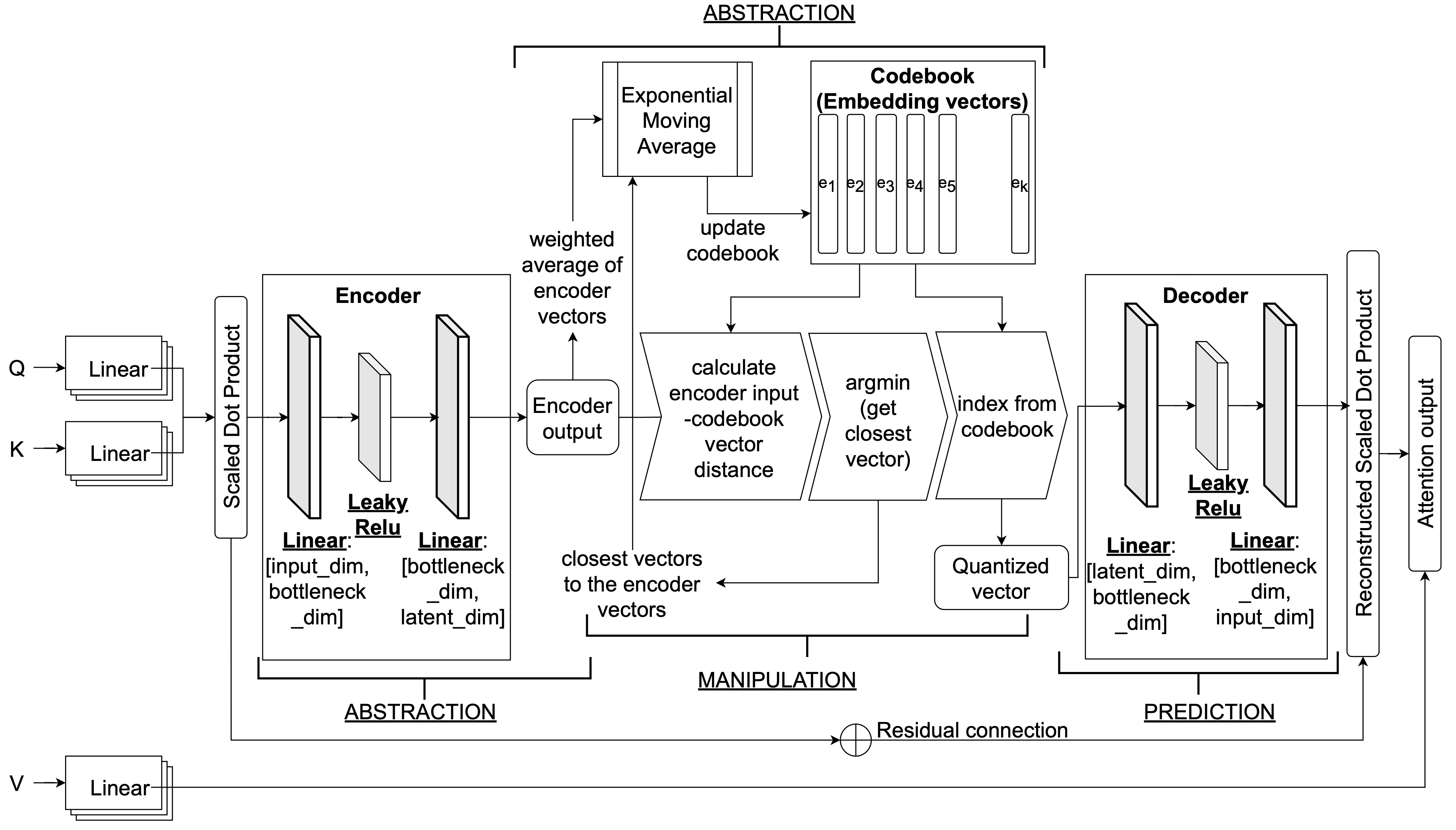} 
    \caption{Proposed architecture of ASAC. The VQVAE module is inseRTEd in the attention mechanism and reconstructs the scaled dot product between queries and keys.}
    \label{fig:asac}
\end{figure}

We show the overall architecture in \autoref{fig:asac}. We plug the attention controller (VQVAE) into the attention mechanism of the transformer, where the attention scores are calculated. The input involves the scaled dot product of the query and keys, which is then fed into the encoder. Both the encoder and the decoder consist of two linear layers separated by a LeakyReLU activation function. A discrete embedding vector matrix, referred to as a codebook, is randomly initialized at the start. The distance between the codebook vectors and the encoding output is calculated, and the closest vectors to the encoder input are indexed from the codebook, and exponential moving average for these vectors is calculated to update the codebook while learning. The chosen vectors from the codebook serve as the input to the decoder. Throughout the training phase, adjustments are made to the codebook to reduce the reconstruction error between the input and output. The output is the reconstructed scaled dot product, via which we compute the final attention. We also add residual connections between the input and the output attentions scores to facilitate stable training. For this, we simply add the reconstructed scaled dot product to the original input scaled dot product.

We employ a composite loss function for the model, comprising three distinct types of losses. The reconstruction loss $L_{\text{recon}}$, is the mean squared error loss between the reconstructed attention and the original attention. The reconstruction loss is defined by the following equation:
\begin{equation}
L_{\text{recon}} = \frac{1}{N} \sum_{i=1}^{N} \left( A^{(i)}_{\text{original}} - A^{(i)}_{\text{reconstructed}} \right)^{2}
\end{equation}

where, $A^{(i)}_{\text{original}}$ and $A^{(i)}_{\text{reconstructed}}$ are the original and reconstructed attention matrix, $N$ is the total number of scores calculated in one forward-pass. The VQVAE attention controller has a vector quantization loss \citep{dosovitskiy2020image} specific to the VQVAE architecture. It ensures that the latent representations (the encodings) match the closest vector in a learnable discrete codebook. The VQ loss can be broken down into two components: the codebook loss and the commitment loss, which help maintain a balance between the flexibility of the codebook and the stability of the encodings. The VQ loss is defined as:

\begin{equation}
L_{VQ} = \| \text{sg}[z_{e}(x)] - e \|_{2}^{2} 
       + \beta \| z_{e}(x) - \text{sg}[e] \|_{2}^{2}
\end{equation}

where, the first term is the codebook loss and the second term is the commitment loss \citep{dosovitskiy2020image}. The codebook loss ensures that the chosen codebook vectors (the embeddings ($e$)) are close to the encoder outputs ($z_{e}(x)$)). ($sg$) represents the stop-gradient operator, meaning no gradient flows through whatever it’s applied on. This loss moves the embedding vectors ($e$) towards the encoder output ($z_{e}(x)$)). The commitment loss encourages the encoder outputs to commit to a codebook vector, preventing them from drifting too far during training, ($\beta$) is a hyper-parameter that balances the importance of the commitment loss relative to the codebook loss. For experiments in this paper, $\beta$ is set to 1.

The task loss, $L_{\text{task}}$ , as the name suggests is the overall loss function specific to the task being solved by the model. It can be binary cross-entropy loss or multi-class cross-entropy based on the task. The cross-entropy loss for multi-class classification is given by:

\begin{equation}
L_{\text{task}} = - \sum_{c=1}^{M} y_{i,c} \, \log(p_{i,c})
\end{equation}

where, ($M$) is the number of classes, ($y_{i,c}$) is a binary indicator of whether class ($c$) is the correct classification for observation ($i$), and ($p_{i,c}$) is the predicted probability that observation ($i$) is of class ($c$). Similarly, if the task is binary classification, then the loss is given as:

\begin{equation}
L_{\text{task}} = - \left[ y_i \log(p_i) + (1 - y_i) \log(1 - p_i) \right]
\end{equation}

where, for the ($i^{th}$) sample, ($y_i$) is the true label (0 or 1), and ($p_i$) is the predicted probability of the
observation being in class 1.

The overall loss is calculated as:
\begin{equation}
L_{\text{final}} = L_{\text{task}} + \lambda \left( L_{\text{recon}} + L_{VQ} \right)
\end{equation}

\subsection{Attention Control on Multi-task Scenario}
\label{subsect:attn_multitask}
The Attention Schema Theory (AST) suggests that cognitive flexibility is essential for smooth transitions between tasks within the human brain. Similarly, a model based on AST should be capable of dynamically transitioning between tasks by adjusting the allocation of attention according to the requirements of each task. A core part of our study involves evaluating our model in scenarios involving binary or multiple tasks, which can range from simple binary or multi-class classification to more complex multi-class multi-label classification.

To perform this experiment, we incorporate unique task identifiers (Task IDs) within the dataset to differentiate between various tasks. Although the dataset remains constant, the corresponding labels may vary depending on the task identifiers. During the classification trials involving an attention controller, the original input images are transformed into a sequence of distinct patches. These patches are then conve\textsf{RTE}d into patch embeddings, with positional embeddings computed to form the final input sequence fed into the model. In a multi-task setting, we augment this input sequence with a task embedding before processing it through the model. We tried the following variations for task information processing:
\begin{itemize}
    \item Including the Task ID information in the input sequence: This is like priming the attention schema with the task to be solved from the beginning.
    \item Including the Task ID information to the VQVAE decoder: This concept is aligned with the notion that the attention schema is adaptable based on the task requirements during the interpretative phase.
    \item Including the Task ID information to both the input sequence and the decoder: This represents holistic integration where attention schema is continuously informed and adjusted based on the task information from initial processing to final decision-making.
\end{itemize}

\subsection{Attention Control on Noisy and Out-of-Distribution Datasets}
\label{subsect:attn_noisyood}
The attention schema enables prioritizing and focusing on the most relevant information. We hypothesize that the AST-based model should also be able to generate this selective attention on data features to attenuate irrelevant or misleading signals. Consequently, the attention schema should enhance the generalization power of the model. To validate this proposition, we assess our model using datasets containing noise and data points outside the distribution while being trained on the original, non-noisy dataset.

\subsection{Assessing Versatility and Adaptability of the Attention Controller}
\label{subsect:versatilty}
The attention schema flexibly adapts to diverse inputs, and due to the discretization of information using a codebook, it enhances resilience to subtle changes in the inputs, such as adversarial attacks. ASAC should also be adept at learning faster with fewer samples by optimizing attention and focusing on relevant information. Additionally, it should facilitate the transfer of knowledge from one task to another by adapting attention patterns. We conduct preliminary investigation on versatility and adaptability of ASAC.

\subsection{Attention Control for Optimizing Resource Distribution}
\label{subsect:resource}
AST posits that the brain creates a simplified model of its attention system to manage cognitive resources efficiently \citep{graziano2015attention}. This internal schema helps predict and adjust attention for various tasks. For instance, when repeatedly performing similar actions like picking up objects, the brain reuses and refines this map to distribute resources efficiently. Conversely, when faced with new challenges, such as multi-tasking, the brain adapts the schema to meet changing demands. We explore whether the ASAC model develops a similar attention schema to allocate cognitive resources effectively.

\section{Experiments on Vision Datasets}
\label{sect:experiments}
In this section, we describe the experiments we conducted and the datasets used. Each subsection contains
the results for those experiments.

\subsection{Attention Controller Performance on Classification}
\label{subsect:attn_perf_classification}
To evaluate the effectiveness of ASAC on transformers, we conducted classification experiments on various vision datasets. For the vision experiments, we used vision transformers\footnote{\url{https://huggingface.co/docs/transformers/main/en/model_doc/vit}} without the ASAC module as a baseline. We compared this baseline to a model with the ASAC module integrated into all layers. Both models were trained from scratch on the datasets, with hyper-parameters and training conditions detailed in the Appendix.

\subsubsection{Datasets}
\label{subsubsect:datasets}
The vision classification tasks include: 1) binary classification on \textsf{Triangles} and \textsf{Polygons}, 2) multi-class classification on \textsf{FashionMNIST}, \textsf{SVHN}, \textsf{TypefaceMNIST}, \textsf{CIFAR10}, \textsf{CIFAR100}, \textsf{Tiny Imagenet}, and \textsf{Places365}, 3) multi-label classification on \textsf{CelebA}, and 4) VQA on the \textsf{Sort-of-Clevr} dataset. Out of these, \textsf{CIFAR10}, \textsf{CIFAR100}\footnote{\url{https://www.cs.toronto.edu/~kriz/cifar.html}}, \textsf{FashionMNIST}\footnote{\url{https://docs.pytorch.org/vision/stable/generated/torchvision.datasets.FashionMNIST.html}}, \textsf{TypefaceMNIST}\footnote{\url{https://www.kaggle.com/datasets/nimishmagre/tMNIST-typeface-MNIST}}, \textsf{SVHN}\footnote{\url{http://ufldl.stanford.edu/housenumbers/}}, \textsf{Tiny Imagenet} \citep{deng2009imagenet}, \textsf{Places365-Standard} \citep{zhou2017places, zhou2017places365hierarchy} and \textsf{CelebA} \citep{liu2015deep} are standard classification datasets.

\begin{figure}[h]
    \centering
    \begin{subfigure}[b]{0.45\textwidth}
        \centering
        \includegraphics[width=\textwidth]{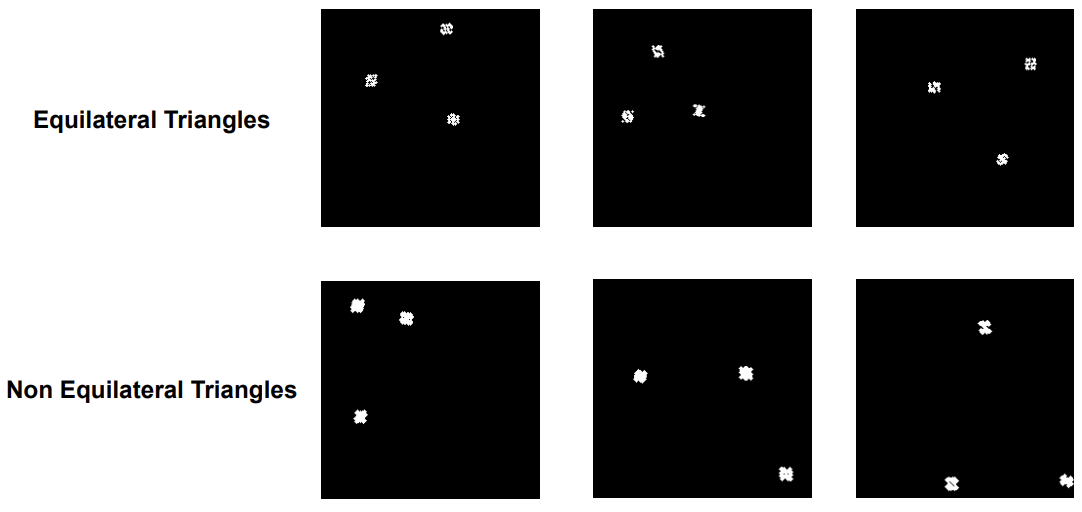}
        \caption{Samples from \textsf{Triangle} dataset.}
        \label{fig:sub1}
    \end{subfigure}
    \hfill
    \begin{subfigure}[b]{0.45\textwidth}
        \centering
        \includegraphics[width=\textwidth]{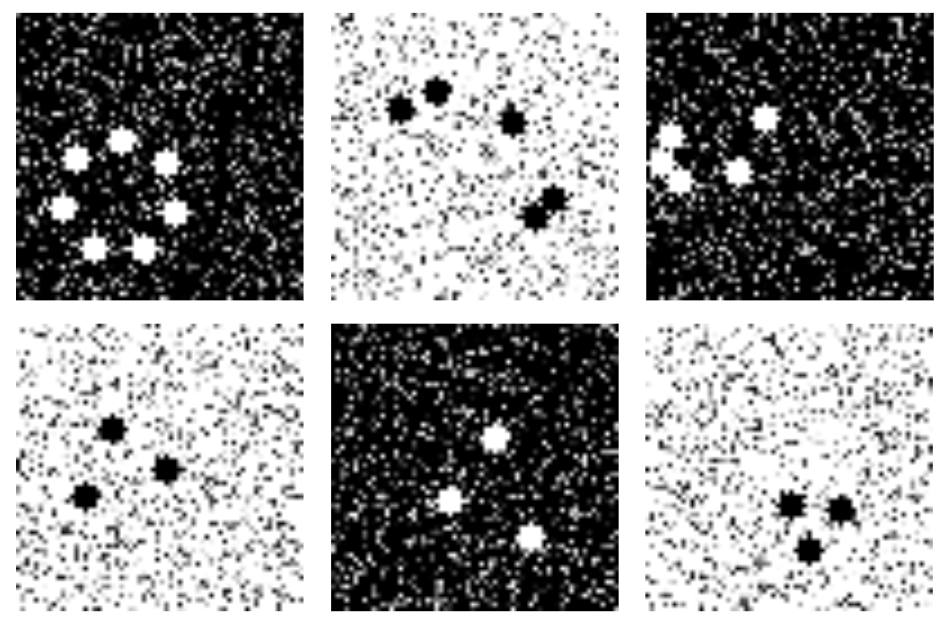}
        \caption{Samples from \textsf{Polygon} dataset.}
        \label{fig:sub2}
    \end{subfigure}
    
    \caption{Samples from different datasets.}
    \label{fig:Triangles_Polygons}
\end{figure}

We explain the rest of the datasets below:
\begin{itemize}
    \item \textsf{Triangles}: We adopt this dataset from \citep{goyal2021coordination, ahmad2022equilateral}, containing images of three white dot clusters on a black background. The task is to predict if the clusters form an equilateral Triangle.
    \item \textsf{Polygons}: This variation of the \textsf{Triangles} dataset includes images with three or more clusters. The task is to identify regular \textsf{Polygons} with equal side lengths, similar to equilateral \textsf{Triangles} but with added complexities like background noise and alternating hues. We train the model on images with 3, 4, or 8 vertices and 5\% noise, and test on images with 5, 6, or 7 vertices and 25\% noise. Images may also be negative, with switched foreground and background colors. We show the samples from \textsf{Triangles} and \textsf{Polygons} data in \autoref{fig:Triangles_Polygons}.
    \item \textsf{Sort-of-Clevr}: \textsf{Sort-of-Clevr} (\textsf{SOC})\footnote{\url{https://github.com/RishikMani/Sort-of-Clevr}} \citep{goyal2021coordination, santoro2017simple} is a dataset for testing reasoning abilities, similar to \textsf{CLEVR} \citep{johnson2017clevr}. It includes images of 2D objects and questions about their properties and relationships. The dataset has two tasks: non-relational reasoning [\textsf{SOC (norel)}], focusing on a single object’s properties like shape and location, and relational reasoning [\textsf{SOC (rel)}], involving relationships between objects like proximity or counting similar shapes.
\end{itemize}
More information on datasets is available in Appendix.

\subsubsection{Results}
\label{subsubsect:results1}

\begin{figure}[ht] 
    \centering
    \includegraphics[width=0.95\textwidth]{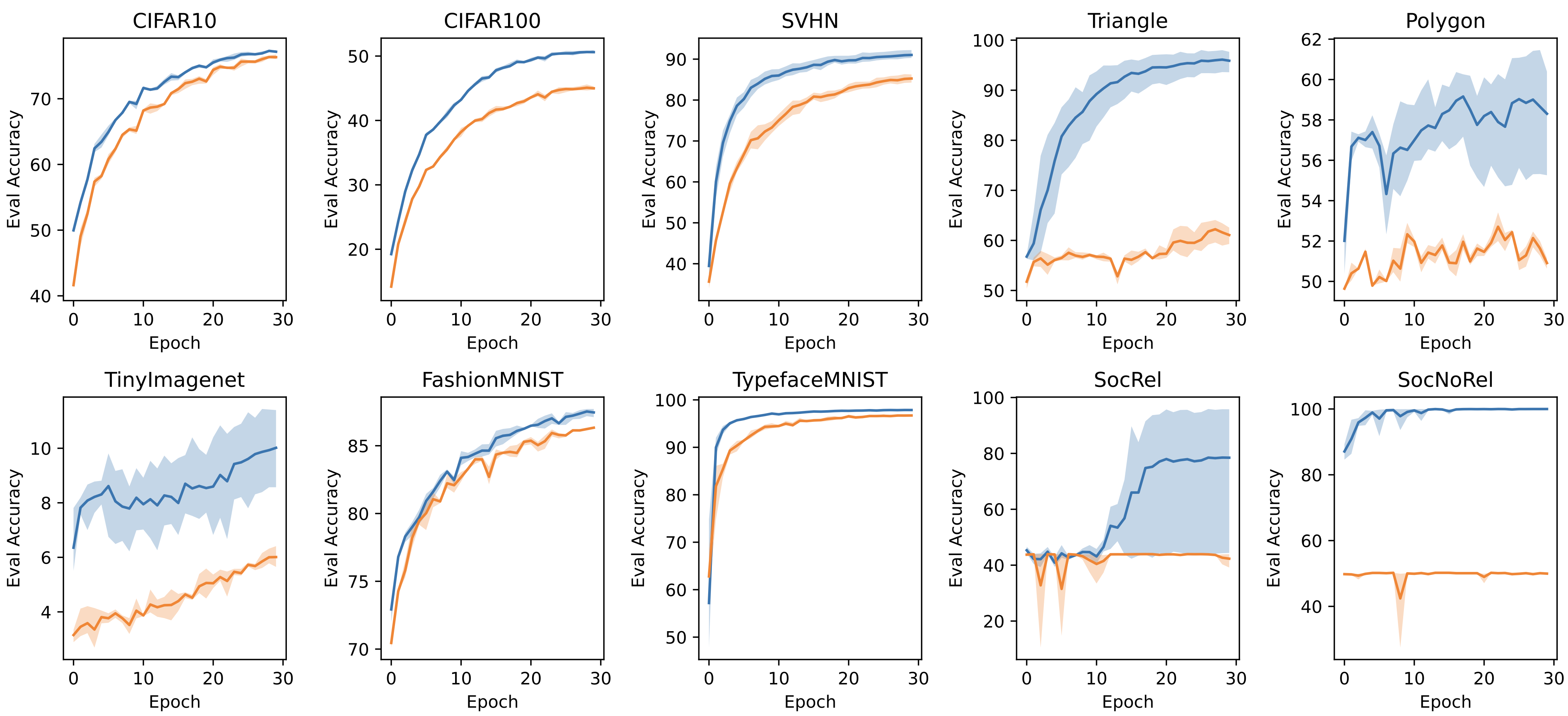} 
    \caption{Classification results on vision datasets. Blue and orange represents ASAC and baseline, respectively.}
    \label{fig:classification1}
\end{figure}

\begin{figure}[ht] 
    \centering
    \includegraphics[width=0.95\textwidth]{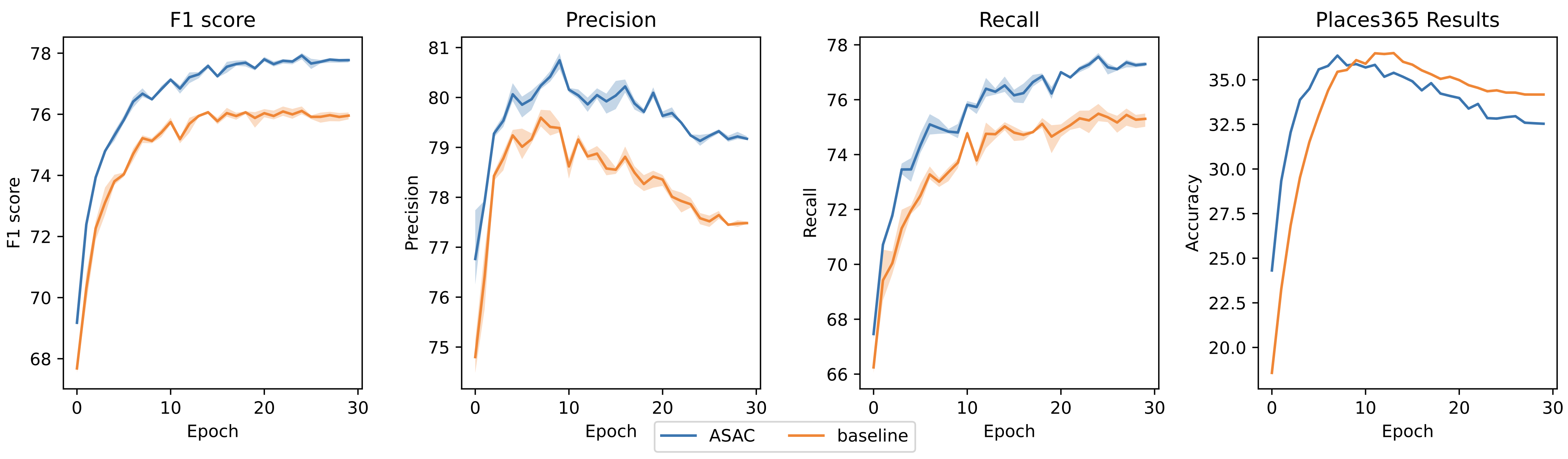} 
    \caption{Classification results on \textsf{CelebA} and \textsf{Places365}.}
    \label{fig:classification2}
\end{figure}

We present the performance of the ASAC architecture in vision classification experiments, as illustrated in \autoref{fig:classification1} and \autoref{fig:classification2}. For all vision experiments in this paper, we trained the model from scratch, without any pre-training, to assess the effectiveness of the attention controller in feature extraction and task understanding across three different seeds, and we aggregated the results. The classification results for all datasets, except \textsf{CelebA} and \textsf{Places365}, are shown in \autoref{fig:classification1}. The results indicate that ASAC outperforms baseline transformers in all experiments and facilitates faster learning compared to the baseline transformers. 

\autoref{fig:classification2} (first three plots) displays the results for the multi-label classification experiment on the \textsf{CelebA} dataset, demonstrating that ASAC achieves superior performance compared to the baseline. The last plot of \autoref{fig:classification2} shows the results for the \textsf{Places365} dataset. Here, ASAC exhibits faster learning and reaches higher accuracy quicker than the baseline. However, after several epochs, we observe over-fitting in both ASAC and the baseline models. Although we show results for 30 epochs for consistency, early stopping would have halted training within 10 epochs for \textsf{Places365} to prevent over-fitting. \textsf{Places365} is a large dataset requiring significant computational resources, and due to cost constraints, we could not run multiple seeds or test various hyper-parameter settings.

\subsection{Multi-task Handling}
\label{subsect:multitask}
In these experiments, we assess the efficacy of the attention controller in comprehending various tasks and
generating predictions according to the task at hand.

\begin{figure}[ht] 
    \centering
    \includegraphics[width=0.5\textwidth]{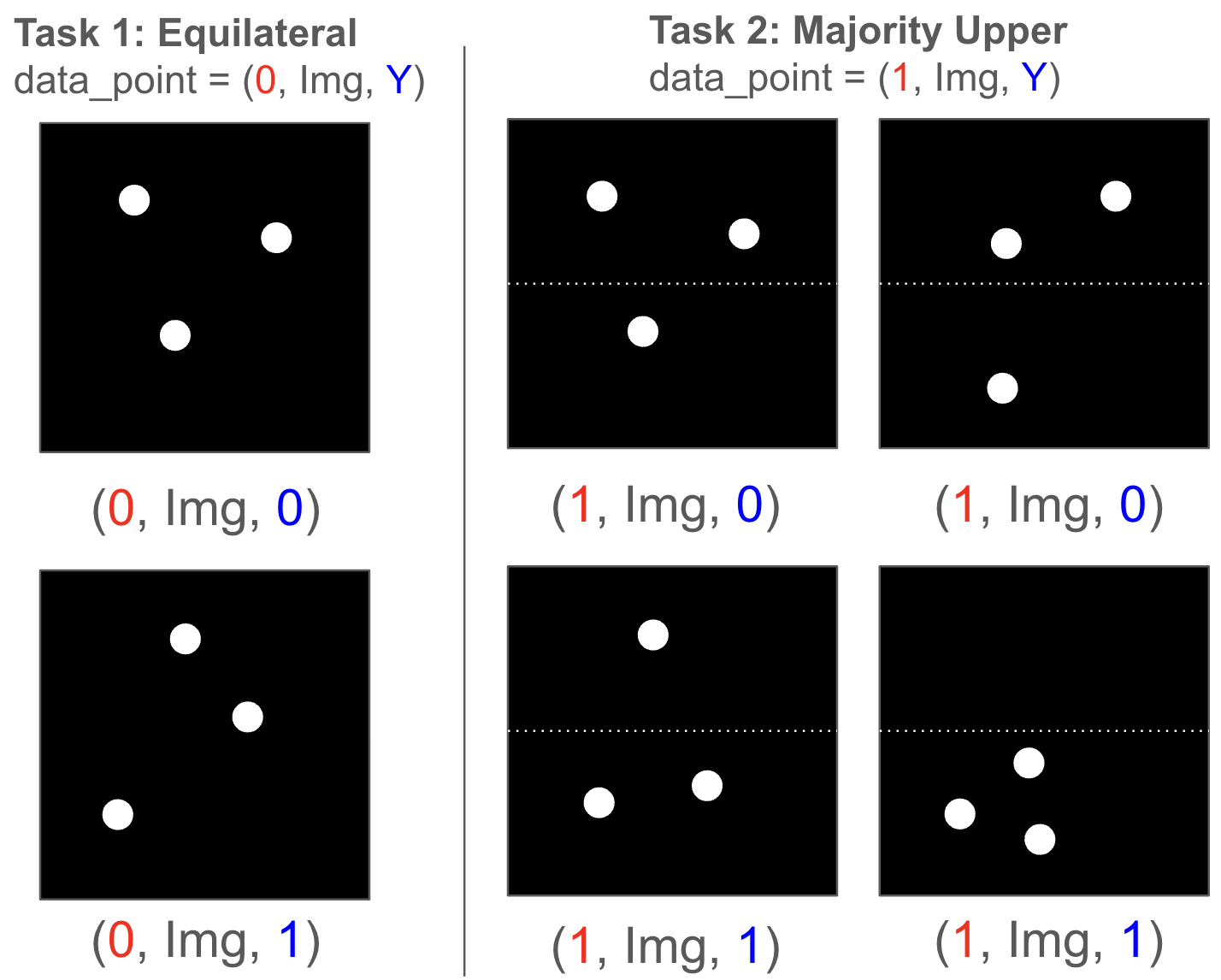} 
    \caption{Multi-task scenario with \textsf{Triangles} data. Left side shows task 1: detecting equilateral \textsf{Triangles} and right side shows task 2: detecting majority upper points. First entry in the data points are the task ID.}
    \label{fig:multitask}
\end{figure}

\subsubsection{Datasets}
\label{subsubsect:datasets2}
We asses the efficacy of the attention controller in comprehending various tasks using the following datasets:
\begin{itemize}
    \item \textsf{Sort-of-Clevr}: We use the same dataset as before, posing relational and non-relational tasks separately.
    \item \textsf{Triangles}: Using the same images as before, we create labels for two tasks: identifying equilateral \textsf{Triangles} and determining if most cluster points are in the upper half. We show a sample in \autoref{fig:multitask}.
    \item \textsf{MNIST}: \textsf{MNIST}\footnote{\url{https://huggingface.co/datasets/ylecun/MNIST}} is classic ML dataset for which we create labels for two tasks: distinguishing odd from even numbers and identifying prime versus composite numbers.
    \item \textsf{Ocular Disease Intelligent Recognition}: \textsf{ODIR-5K}\footnote{\url{https://www.kaggle.com/datasets/andrewmvd/ocular-disease-recognition-odir5k}} \citep{prabhakaran2024enhance} is real-world ophthalmic disease dataset with 8 labels. We pose it as a multi-task problem, with each task being a binary classification of whether an image represents a specific disease.
    \item \textsf{Places365}: Using the same images, we create three tasks: single-label classification, and multi-label classification into level-1 and level-2 hierarchy labels \citep{zhou2017places, zhou2017places365hierarchy}.
    \item \textsf{CIFAR10}-\textsf{SVHN}: We mix the \textsf{CIFAR10} and \textsf{SVHN} dataset and the task is to label the images of \textsf{CIFAR10} or \textsf{SVHN} based on the task ID input.
\end{itemize}

More information on the datasets is available in Appendix.

\subsubsection{Results}
\label{subsubsect:results_multitask}
\autoref{tab:table_multi} illustrates that the AST-based attention controller exhibits superior capability in adjusting the attention schema within a multi-task environment when compared to the model lacking the controller. In these experiments, we run three different seeds and present aggregate results. We see that the best-performing model among different datasets is a different variant of the model. In datasets like \textsf{ODIR-5K}, \textsf{Triangles} and \textsf{CIFAR10}-\textsf{SVHN}, the taskID in input performs the best, which means task-specific information (TaskID) provided at the input level helps the model to quickly adapt and focus on the relevant features for each task. We observe that task information in the decoder helps in \textsf{Sort-of-Clevr}. By sending the TaskID to the decoder, the model can leverage the task-specific information at a later stage, allowing the encoder to focus on extracting general features without being biased by the task. For \textsf{MNIST}, we see that the best performing model is when the task information is present in the input as well as the decoder, meaning that the model is able to make better decisions when the model is continuously informed and adjusted based on the task information. Since \textsf{Places365} is a huge dataset and the tasks vary significantly (single-label and multi-label classification in different tasks), we assumed that task information in both the input and the decoder will help solve this task as the model can retrieve features based on the task and also conditioned to predict the final output based on the task information. We see ASAC performing better than the baseline in this setting for \textsf{Places365}.

\begin{table}[]
\centering
\caption{Results of the multi-task experiments. Due to cost constraints for computational resources, we run \textsf{Places365} with only one variant where the taskID is added to the VQVAE input and decoder. Bold values represent the best performance.}
\label{tab:table_multi}
\resizebox{0.9\textwidth}{!}{%
\begin{tabular}{l|l|l|lll}
                          & Model $\rightarrow$     & Baseline    & \multicolumn{3}{l}{ASAC}                                                                                    \\ \cline{2-6} 
                          & TaskID in $\rightarrow$ & input       & \multicolumn{1}{l|}{input}                & \multicolumn{1}{l|}{decoder}             & input\&decoder       \\ \cline{2-6} 
Metrics $\downarrow$                   & Dataset $\downarrow$                   &             & \multicolumn{1}{l|}{}                     & \multicolumn{1}{l|}{}                    &                      \\ \hline
\multirow{5}{*}{Accuracy} & \textsf{SOC}                       & 63.15 $\pm$ 7.39 & \multicolumn{1}{l|}{67.08 $\pm$ 2.84}          & \multicolumn{1}{l|}{\textbf{73.23 $\pm$ 6.4}} & 71.09 $\pm$ 4.35          \\
                          & \textsf{MNIST}                     & 96.78 $\pm$ 0.4  & \multicolumn{1}{l|}{97.17 $\pm$ 0.13}          & \multicolumn{1}{l|}{97.04 $\pm$ 0.26}         & \textbf{97.35 $\pm$ 0.32} \\
                          & \textsf{CIFAR10}-\textsf{SVHN}              & 85.94 $\pm$ 1.72 & \multicolumn{1}{l|}{\textbf{86.58 $\pm$ 0.64}} & \multicolumn{1}{l|}{84.25 $\pm$ 0.28}         & 86.56 $\pm$ 0.63          \\
                          & \textsf{Triangles}                 & 78.66 $\pm$ 0.83 & \multicolumn{1}{l|}{\textbf{96.71 $\pm$ 1.28}} & \multicolumn{1}{l|}{96.01 $\pm$ 1.39}         & 94.39 $\pm$ 3.2           \\
                          & \textsf{ODIR-5K}                   & 83.44 $\pm$ 0.01 & \multicolumn{1}{l|}{\textbf{83.46 $\pm$ 0.01}} & \multicolumn{1}{l|}{75.60 $\pm$ 5.28}         & 79.32 $\pm$ 5.24          \\ \hline
Precision                 & \textsf{Places365}                 & 60.46       & \multicolumn{1}{l|}{-}                    & \multicolumn{1}{l|}{-}                   & \textbf{61.03}       \\
Recall                    &                           & 54.04       & \multicolumn{1}{l|}{-}                    & \multicolumn{1}{l|}{-}                   & \textbf{54.56}       \\
F1                        &                           & 56.25       & \multicolumn{1}{l|}{-}                    & \multicolumn{1}{l|}{-}                   & \textbf{56.82}      
\end{tabular}%
}
\end{table}

\subsection{Generalization Power}
\label{subsect:generalization}
In these experiments, we train the model with non-noisy original datasets and test on noisy and out-of-distribution (OOD) datasets to check the generalization power of the attention controller.

\subsubsection{Datasets}
\label{subsubsect:dataset3}
We test the generalization power of our model two noisy datasets: \textsf{CIFAR-10C}\footnote{\url{https://github.com/hendrycks/robustness}} and \textsf{Tiny Imagenet}-C\footnote{\url{https://github.com/hendrycks/robustness}}. Additionally, we test ASAC on the following OOD datasets:
\begin{itemize}
    \item \textsf{Triangles-OOD}: We create test images with different sizes of clusters. We also vary the shape of clusters to circles, squares, or Triangles and the shape can be empty or filled. We train using the original (filled, circle, constant size clusters) \textsf{Triangles} dataset.
    \item \textsf{Polygons-OOD}: The train set contains 3,4,8 vertices with colored noise on 5\% of the background. The test set contains 5, 6, and 7 vertices with colored noise on 25\% of the background area.
\end{itemize}

\subsubsection{Results}
\label{subsubsect: result_generalization}
\begin{figure}[ht] 
    \centering
    \includegraphics[width=0.65\textwidth]{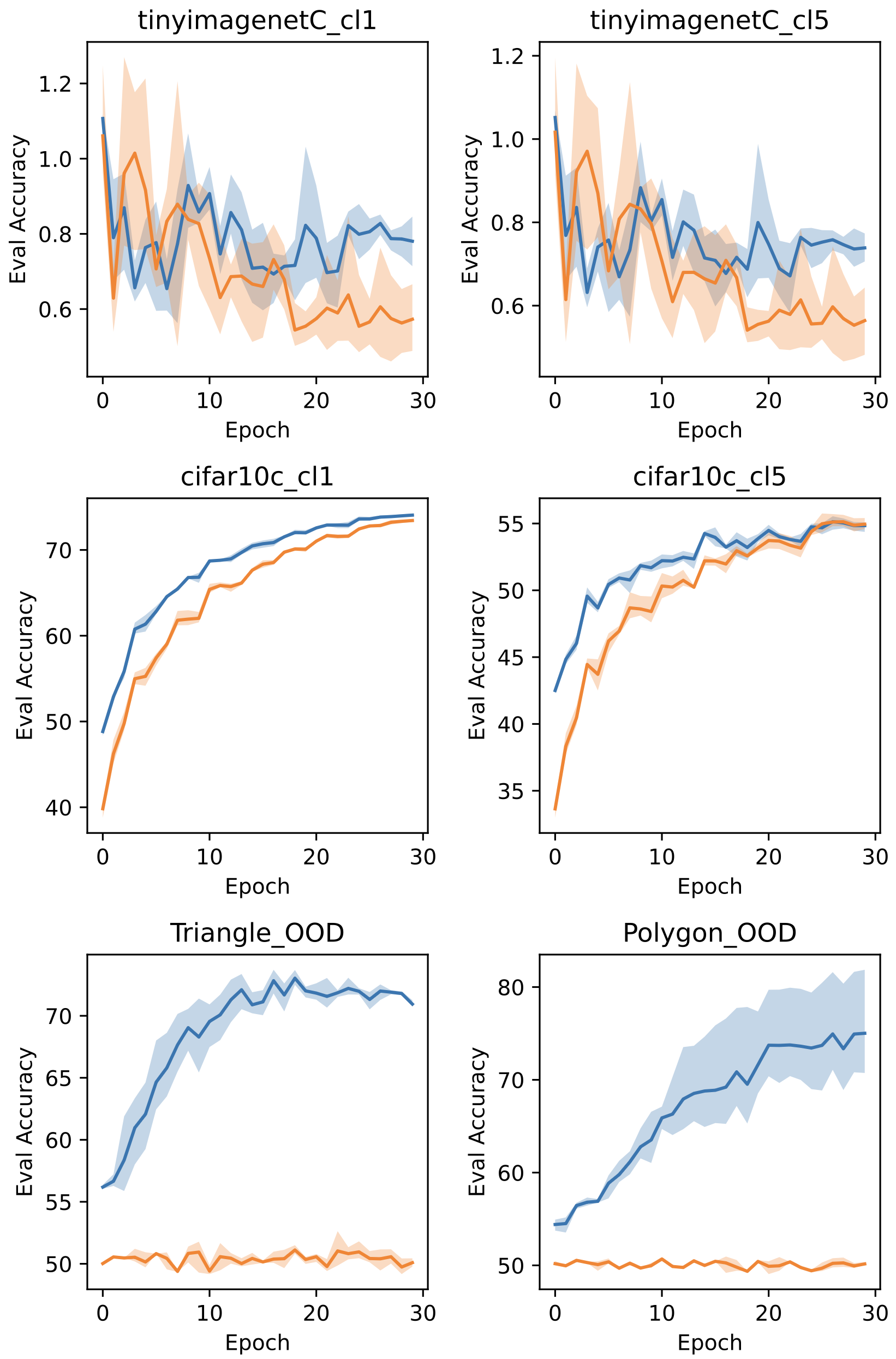} 
    \caption{Generalization results on noisy and OOD datasets. First and second row are the results of corrupted TinyImagenet and \textsf{CIFAR10}, respectively. Last row shows the results for OOD experiments for \textsf{Triangles} and \textsf{Polygons}. Blue and orange color represnts ASAC and baseline, respectively.}
    \label{fig:generalization}
\end{figure}

The results for generalization experiments are shown in \autoref{fig:generalization}. There is a marginal increase in performance with attention controller on corrupted \textsf{Tiny Imagenet} dataset. Since this is a challenging dataset, the overall accuracy obtained by both the baseline and VQVAE models is relatively low. The ASAC model achieves higher accuracies in corrupted CIFAR datasets. Furthermore, we also see significant improvement in the OOD experiments with \textsf{Triangles} and \textsf{Polygons} data.

\subsection{Versatility and Adaptability}
\label{subsect:versatility}
We conduct preliminary investigations to evaluate the versatility and adaptability of the model with the attention controller. Our experiments focus on four tasks: adversarial attacks, transfer learning, few-shot learning, and learning efficiency. 

For the adversarial attacks experiments, we train the model for 20 epochs on the original dataset and test it on perturbed images. These attacked images, generated using the Fast Gradient Sign Method (FGSM) \citep{goodfellow2014explaining} and Projected Gradient Descent Method (PGDM) \citep{ayas2022projected}, are slightly altered to cause the model to misclassify them. FGSM calculates the gradient of the model’s loss with respect to the image and adds a scaled version of this gradient to the original image, where the scale factor, $\epsilon$, controls the perturbation magnitude. PGDM iteratively applies small perturbations, recalculating the gradient at each step to maximize the model’s error.

For transfer learning experiment, we pre-train the model on one dataset for 10 epochs and fine-tune on another (related) dataset for 5 epochs. For the few-shot experiments, we train the model for 10 epochs on one dataset and then use different percentages of another data to do few-shot training. For testing the learning efficiency, we train the model using different percentages of data for 25 epochs and test on the original test set of the same dataset. We run the experiments for three different seeds and show the aggregated results. For these experiments, we used \textsf{CIFAR10}, \textsf{CIFAR100}, \textsf{FashionMNIST}, \textsf{SVHN} dataset, described in \autoref{subsubsect:datasets}.

\begin{figure}[ht] 
    \centering
    \includegraphics[width=0.7\textwidth]{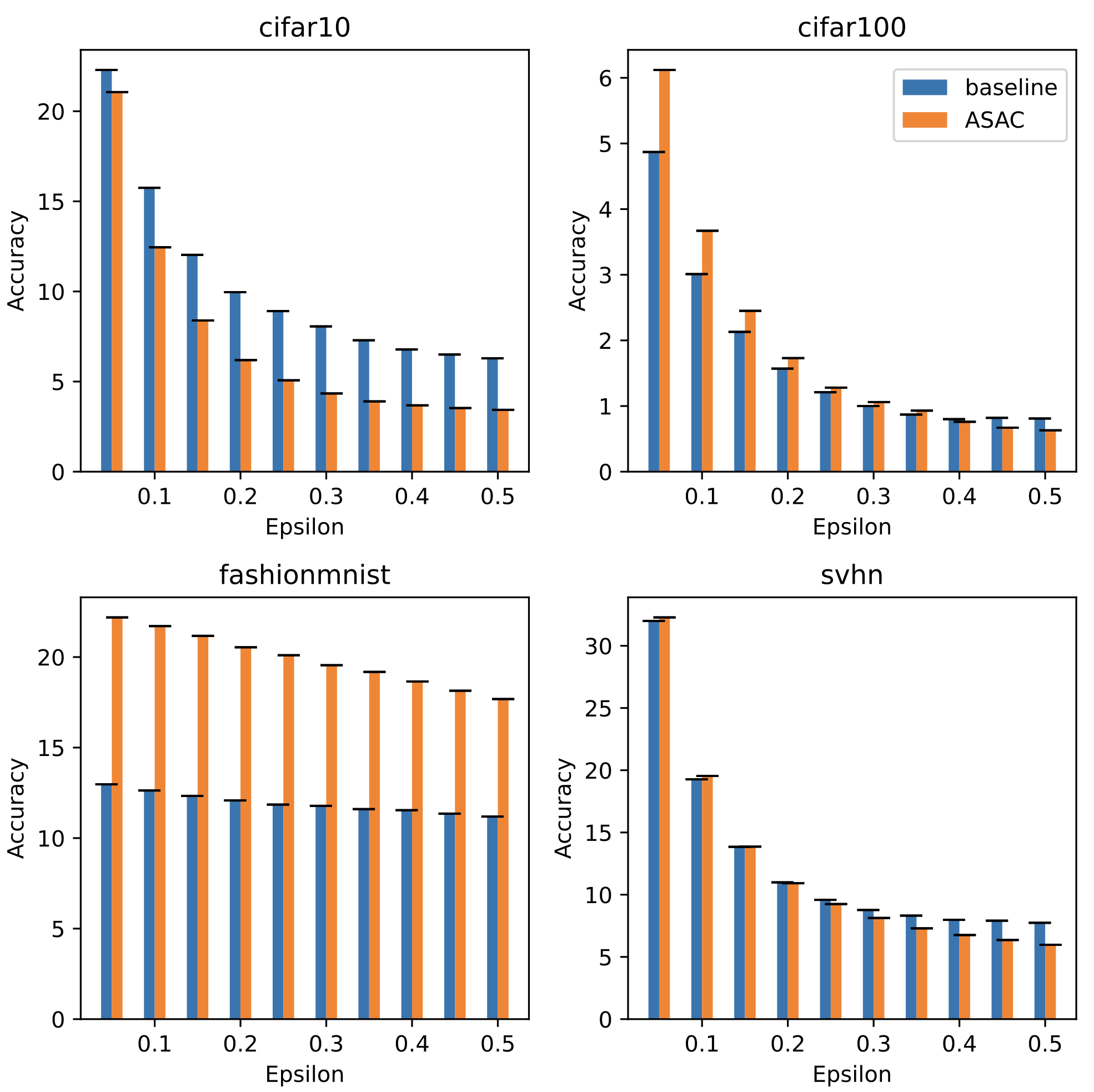} 
    \caption{Adversarial attacks results for FGSM attacks on different datasets.}
    \label{fig:fgsm}
\end{figure}

\begin{figure}[ht] 
    \centering
    \includegraphics[width=0.7\textwidth]{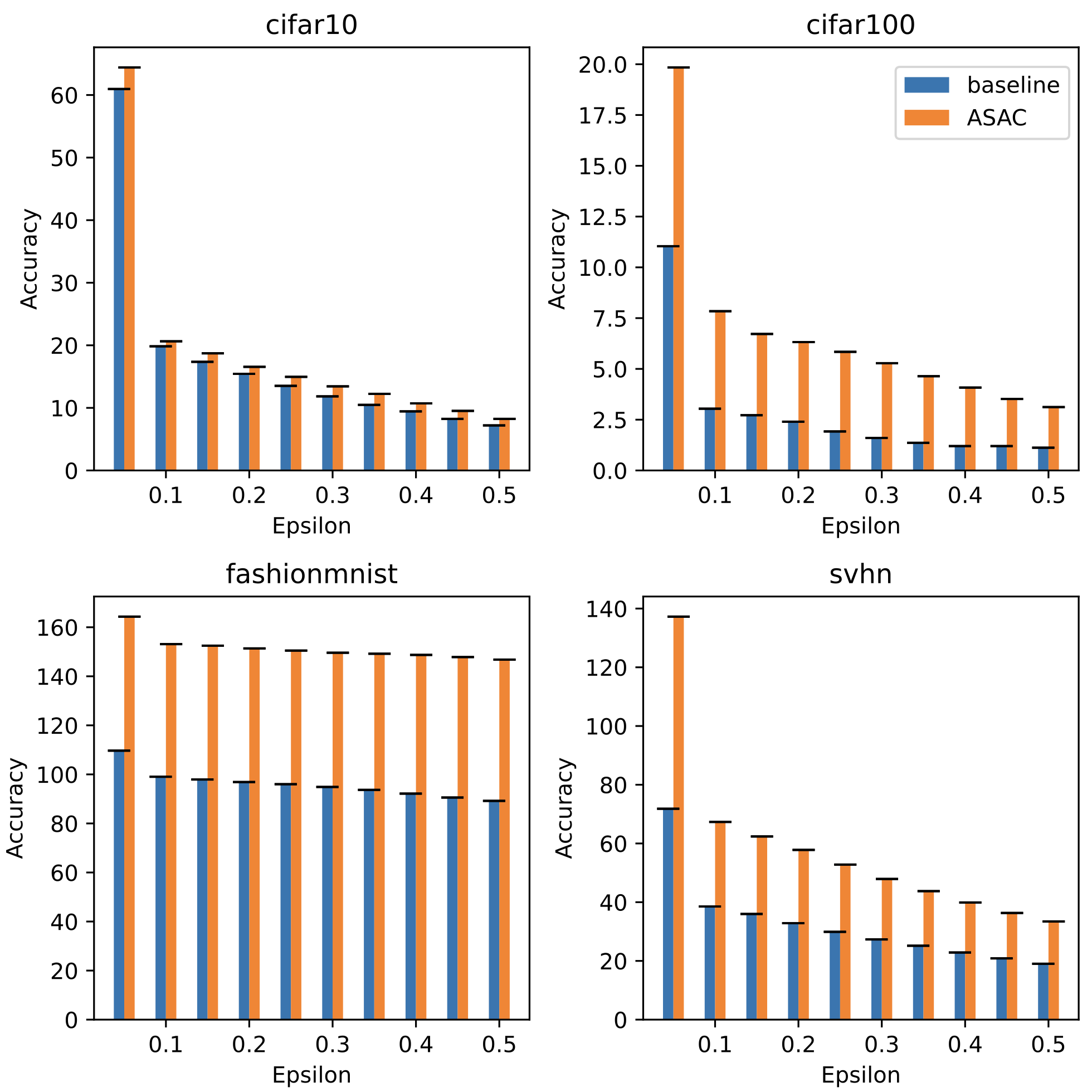} 
    \caption{Adversarial attacks results for PGDM attacks on different datasets.}
    \label{fig:pgdm}
\end{figure}

\subsubsection{Results}
\label{subsubsect:verstality_results}
We show the results for FGSM and PGDM adversarial attacks in \autoref{fig:fgsm} and \autoref{fig:pgdm}. In the FGSM adversarial attacks, the performance varies. Specifically, for \textsf{CIFAR10}, ASAC consistently underperforms compared to the baseline across all $\epsilon$ values. Conversely, ASAC demonstrates superior performance for \textsf{FashionMNIST} across all $\epsilon$ values. For \textsf{CIFAR100} and \textsf{SVHN}, ASAC exhibits better performance at smaller $\epsilon$ values; however, as $\epsilon$ values increase, indicating stronger attacks, the baseline begins to outperform ASAC. In contrast, the results for PGDM attacks show that ASAC consistently surpasses the baseline across all datasets and $\epsilon$ values.

The mixed results for FGSM attacks can be attributed to the nature of the datasets and the varying complexity of adversarial examples. For \textsf{CIFAR10}, the simpler architecture of the baseline might be more resilient to minor perturbations, leading to better performance at all $\epsilon$ levels. For \textsf{FashionMNIST}, the inherent robustness of ASAC’s attention mechanisms likely provides it with an advantage over the baseline, enabling it to better handle adversarial examples. In the cases of \textsf{CIFAR100} and \textsf{SVHN}, ASAC’s superior performance at lower $\epsilon$ values could be due to its ability to capture complex features effectively. However, as the attack strength increases, the baseline’s more straightforward architecture might become advantageous by avoiding overfitting to perturbed data, thus outperforming ASAC in higher $\epsilon$ scenarios. Regarding PGDM attacks, ASAC’s consistent outperformance across all datasets and $\epsilon$ values highlights its robust design. The combination of improved feature extraction and attention mechanisms likely equips ASAC with enhanced capabilities to mitigate the effects of stronger, iterative adversarial attacks compared to the baseline.

\begin{figure}[ht] 
    \centering
    \includegraphics[width=0.7\textwidth]{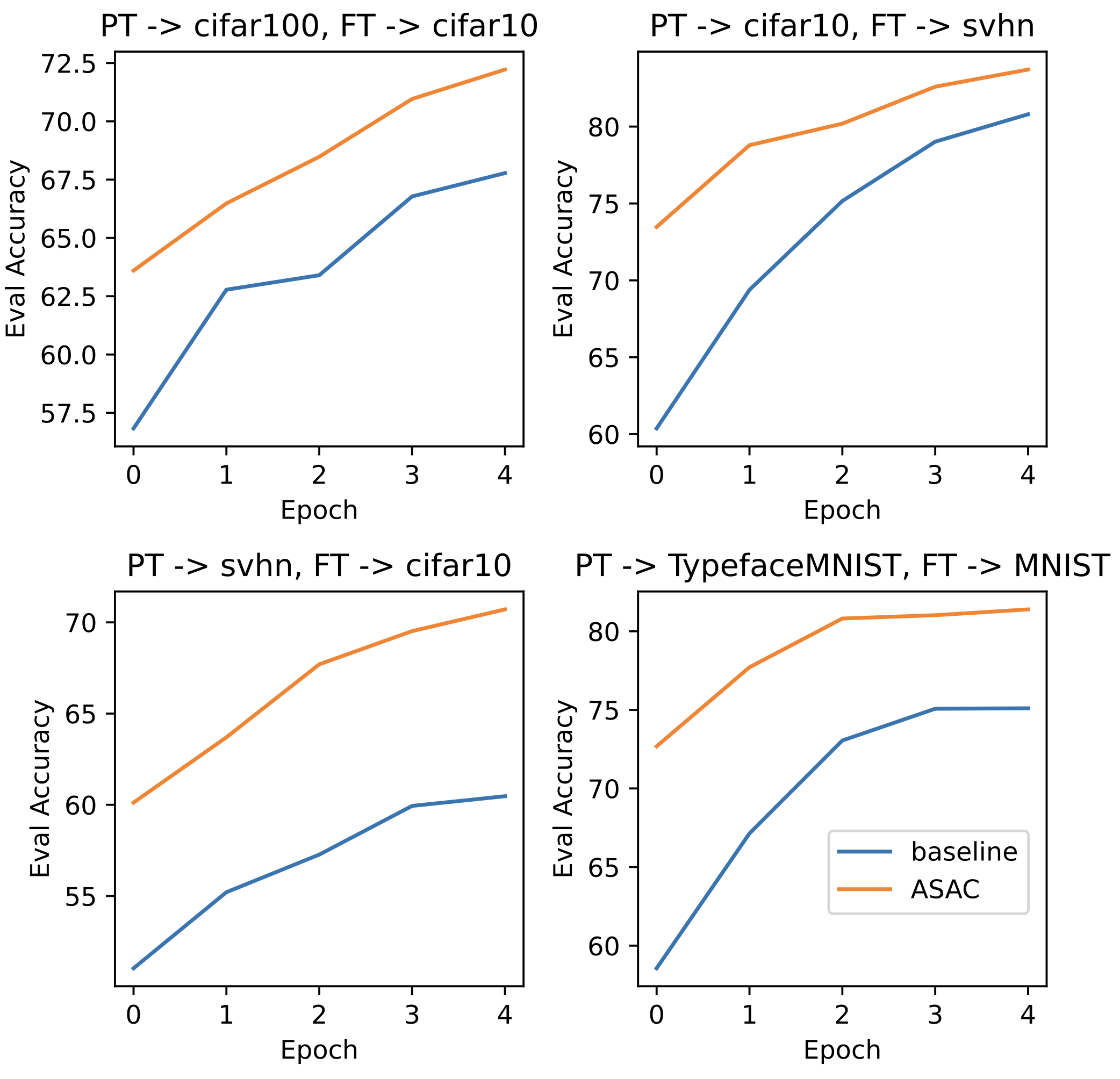} 
    \caption{Results of transfer learning experiments. PT and FT stands for datasets used for pre-training and
fine-tuning.}
    \label{fig:transfer_learning}
\end{figure}

The results of our transfer learning experiments are illustrated in \autoref{fig:transfer_learning}. We observe that ASAC consistently outperforms the baseline when trained on one dataset and fine-tuned on another. This superior performance can be attributed to ASAC’s ability to effectively utilize attention patterns during the transfer of knowledge. Its design allows it to selectively focus on crucial aspects of the pre-training data, which enhances its generalization capability. Consequently, ASAC not only learns robust representations from the initial dataset but also successfully adapts and fine-tunes these representations to excel on the target dataset.

\begin{figure}[ht] 
    \centering
    \includegraphics[width=0.7\textwidth]{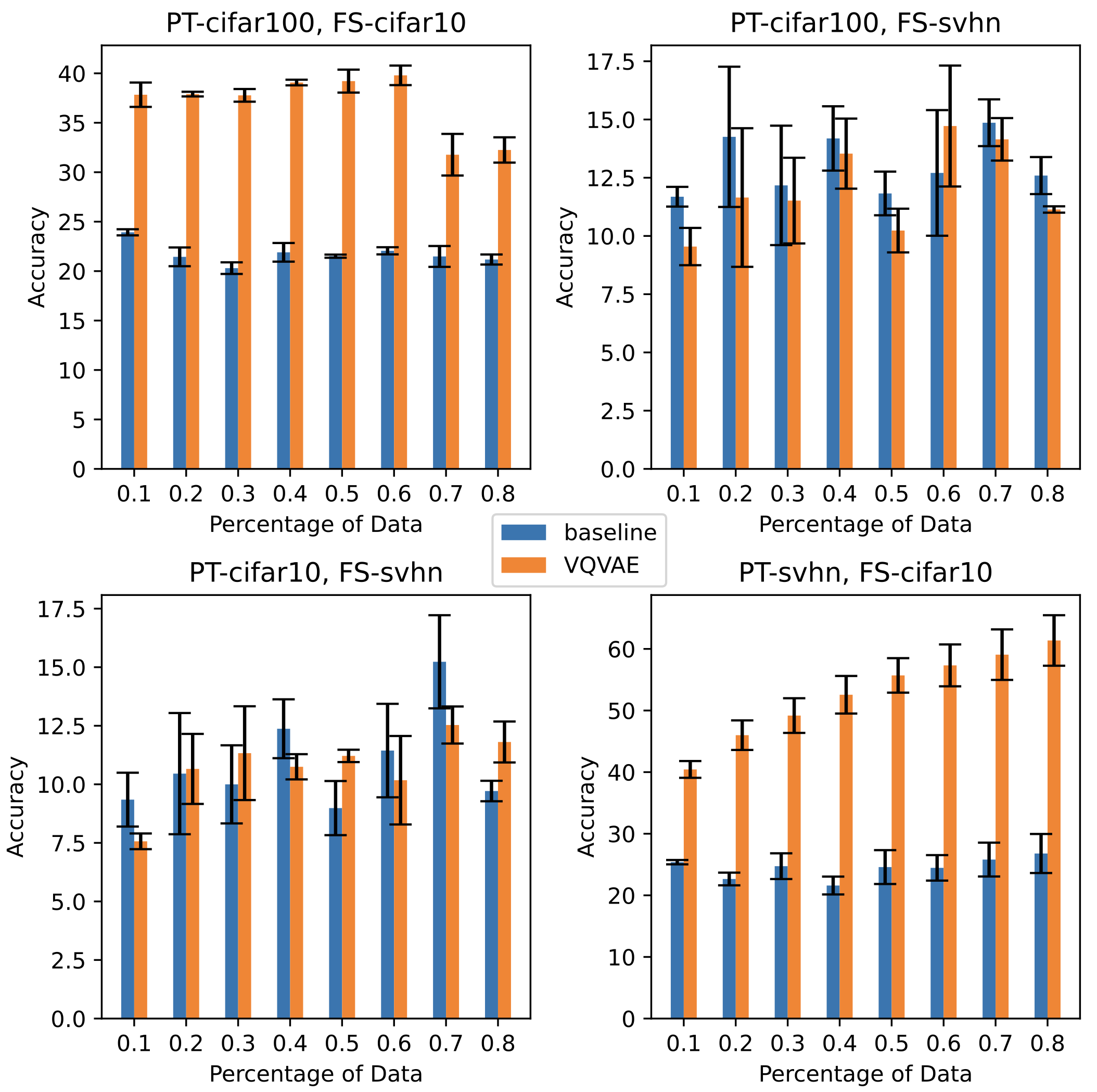} 
    \caption{Results of few-shot experiments. PT and FS stands for datasets used for pre-training and few-shot
fine-tuning.}
    \label{fig:few_shot}
\end{figure}

We show the results of few-shot learning in \autoref{fig:few_shot}. We observe that ASAC demonstrates enhanced accuracies when pre-training is performed on \textsf{CIFAR100} and fine-tuning on \textsf{CIFAR10}, as well as in another experiment where pre-training is conducted on \textsf{SVHN} followed by few-shot fine-tuning on \textsf{CIFAR10}. These improvements can be attributed to the complementary nature of the datasets and the effective transfer of learned features, allowing ASAC to leverage its attention mechanisms efficiently.

For the former combination, the shared object-centric nature of \textsf{CIFAR100} and \textsf{CIFAR10} allows ASAC to apply its learned representations effectively, resulting in improved performance.

In the case of pre-training on \textsf{SVHN} and fine-tuning on \textsf{CIFAR10}, the larger number of training samples in \textsf{SVHN} may contribute to the development of more robust initial features. However, this scenario may not be as straightforward, and the effective transfer could also be attributed to the general adaptability of ASAC’s attention mechanisms, which manage to refine and adapt learned features efficiently to the new context presented by \textsf{CIFAR10}.

Conversely, for the scenarios where pre-training is done on \textsf{CIFAR100} or \textsf{CIFAR10} and few-shot fine-tuning on \textsf{SVHN}, we observe mixed results across different few-shot percentages of the data. This variability can be explained by the differences in data distribution and feature complexity between these datasets. While CIFAR datasets focus more on varied objects, \textsf{SVHN} is centered around digit recognition in different contexts, leading to potential mismatches in learned features during transfer learning. Consequently, the effectiveness of knowledge transfer fluctuates, resulting in inconsistent performance for different percentages of the few-shot data.

\begin{figure}[ht] 
    \centering
    \includegraphics[width=0.75\textwidth]{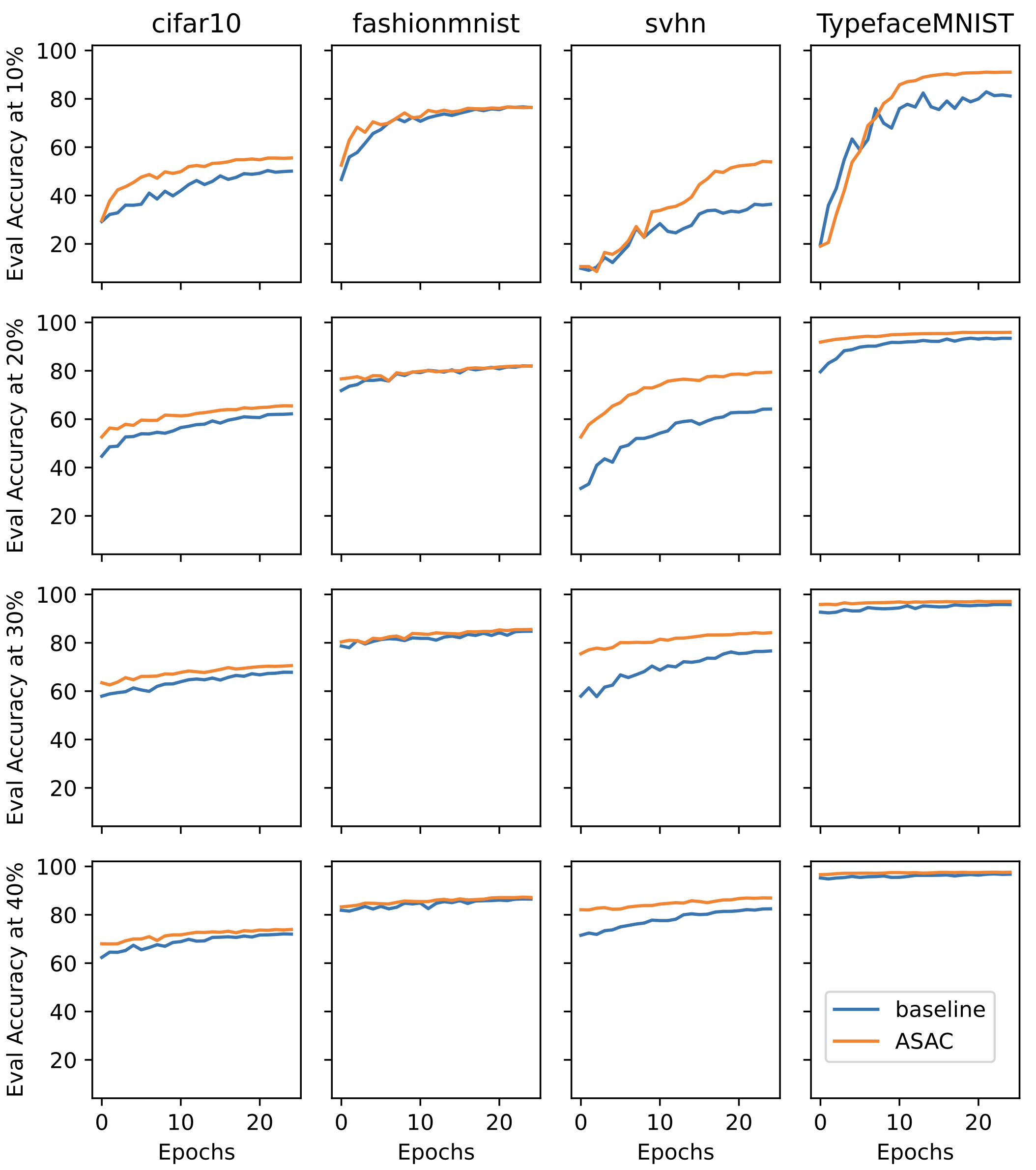} 
    \caption{Results of learning efficiency experiments. Each row represents results for training the model
with different percentages of data.}
    \label{fig:learning_efficiency}
\end{figure}

Our findings on learning efficiency experiments, depicted in \autoref{fig:learning_efficiency}, reveal that ASAC consistently outperforms the baseline across all datasets and varying data percentages used for training the models. ASAC excels in rapid learning with fewer samples by optimizing attention and effectively focusing on relevant information. This efficiency allows ASAC to achieve higher accuracies compared to the baseline, starting from the very initial epochs.

\subsection{Allocation of Cognitive Resources}
\label{subsect:allocation}
In the VQVAE architecture, the codebook is crucial as it represents part of the attention schema, enabling
efficient allocation of computational resources. We test this by analyzing the \textsf{ODIR-5K} and \textsf{CIFAR10}-\textsf{SVHN}
experiments in multi-task scenarios, whose results are presented in \autoref{subsubsect:results_multitask}. Specifically, we calculate and analyze the usage of distinct codes from the codebook for different tasks to demonstrate effective resource allocation.

\subsubsection{Results}
\label{subsubsect:alloc_results}
\begin{figure}[ht] 
    \centering
    \includegraphics[width=0.5\textwidth]{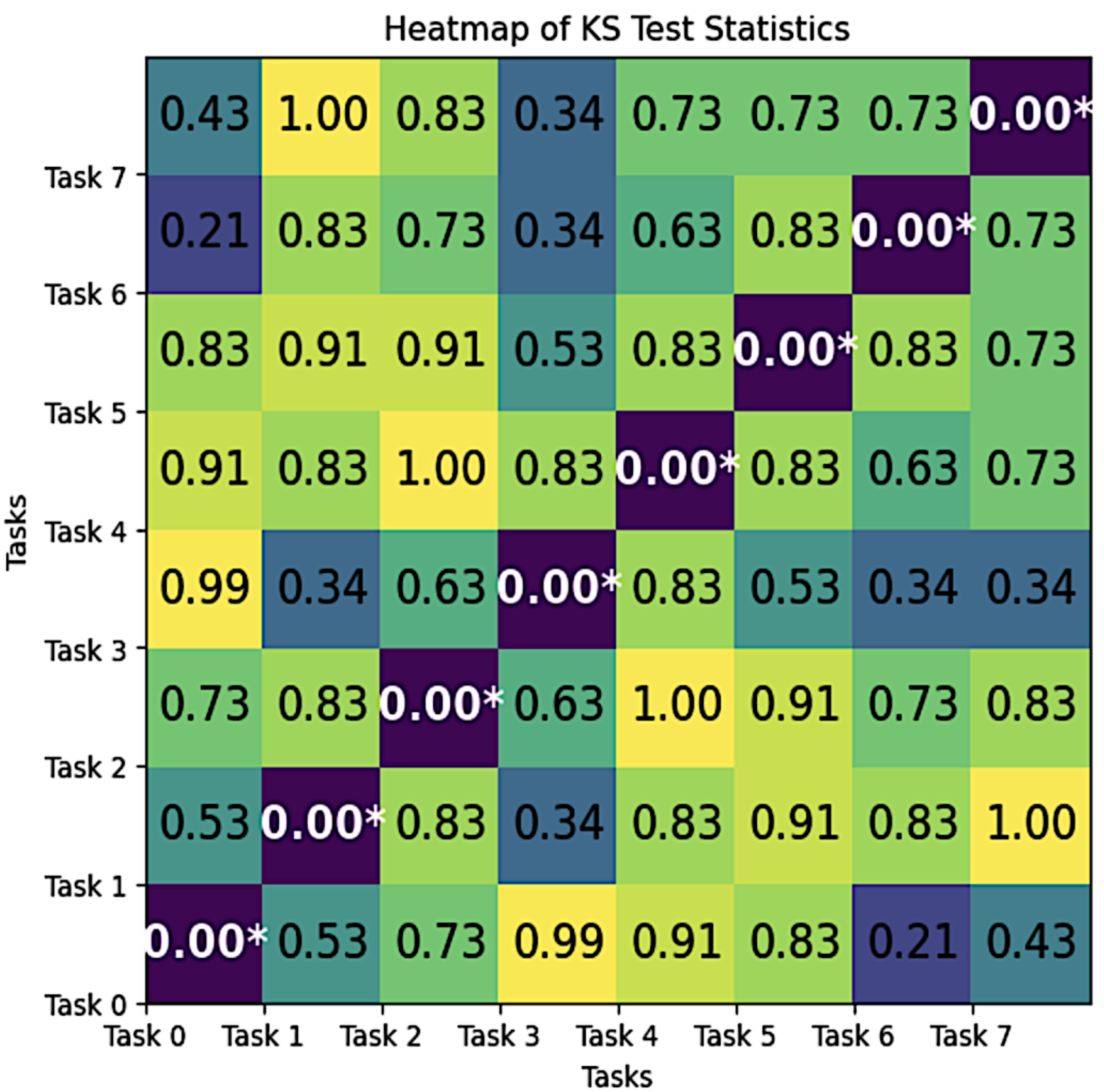} 
    \caption{KS test \textit{p-value} for \textsf{ODIR-5K}.}
    \label{fig:heatmap}
\end{figure}

We assess codebook usage by counting the frequency of specific codes for each task and calculating the Kolmogorov-Smirnov (KS) test \textit{p-value}. For the \textsf{CIFAR10}-\textsf{SVHN} dataset, ASAC produces a \textit{p-value} of 0.0052, indicating significant differences in code usage between tasks. This suggests the model uses distinct codes to label images in \textsf{CIFAR10} (everyday objects) and \textsf{SVHN} (house numbers), given their differing image types. For \textsf{ODIR-5K}, we calculate pairwise KS-Test \textit{p-values} for 8 tasks shown in \autoref{fig:heatmap}. All pairwise \textit{p-values} fall between 0.21 and 1.0, with most above 0.7. Since no value for \textsf{ODIR-5K} is below the significance level of 0.05, the results suggest similar codebook usage across tasks, indicating code reuse for solving different tasks but with similar images.

\section{ASAC Integration to Language Models}
\label{sect:languagemodelexp}
We tested the ASAC integration with a small NLP model using simple classification tasks. Training an NLP model from scratch is impractical due to the vast datasets and significant computational resources required. Thus, we use pre-trained \textsf{DistilBERT} model \citep{sanh2019DistilBERT} for NLP experiments. However, integrating the VQVAE module into all layers of such a pre-trained model presents challenges. Ideally, one would train only the new VQVAE parameters while keeping the pre-trained weights unchanged. However, adding untrained parameters to a pre-trained network disrupts the synergy between components, as the pre-trained weights are already optimized to work together, whereas the new VQVAE parameters have not undergone this collective training process. This misalignment causes the model to struggle to adapt, leading to poor performance.

To prevent disrupting the pre-trained weights, we incorporate the ASAC module solely into the last layer of the model, with its parameters initialized randomly. The baseline model retains pre-trained weights for all layers except the last one, which is initialized randomly. Similarly, the tested model maintains pre-trained weights for all layers but includes the ASAC module in the last layer. We then fine-tune both the baseline \textsf{DistilBERT} model and the ASAC-integrated \textsf{DistilBERT}. We test the model on GLUE\footnote{\url{https://huggingface.co/datasets/nyu-mll/glue}} benchmark datasets.

\subsubsection{Results}
\label{subsubsect:nlp_results}
We show the results with the GLUE dataset in \autoref{tab:glue_results}. We fine-tune the \textsf{DistilBERT} model for five different seeds and present the aggregate result. We observe that for all the experiments, the ASAC model outperforms the baseline. We also report the \textit{p-value} to indicate the statistical significance of the differences between the VQVAE performance and the baseline. A lower \textit{p-value} (typically less than 0.05) suggests that the differences are statistically significant and unlikely to be due to random variation, thereby indicating that the VQVAE integration has a meaningful impact on the model’s performance. However, in some instances, the \textit{p-value} exceeds 0.05, indicating that the results may not be statistically significant.

\begin{table}[ht]
\centering
\caption{Results with GLUE dataset}
\label{tab:glue_results}
\resizebox{0.75\textwidth}{!}{%
\begin{tabular}{l|l|l|l|l}
GLUE Dataset          & Metric    & Baseline       & ASAC                    & p-value         \\ \hline
\multirow{2}{*}{\textsf{STSB}} & Pearson   & 0.8644 $\pm$ 0.0029 & \textbf{0.8651 $\pm$ 0.0012} & 0.5795          \\
                      & SpearmanR & 0.8622 $\pm$ 0.0027 & \textbf{0.8625 $\pm$ 0.0010} & 0.8327          \\
\textsf{SST2}                  & Accuracy  & 0.8986 $\pm$ 0.0102 & \textbf{0.9052 $\pm$ 0.0025} & \textbf{0.0001} \\
\textsf{QNLI}                  & Accuracy  & 0.8850 $\pm$ 0.0019 & \textbf{0.8895 $\pm$ 0.0027} & \textbf{0.0105} \\
\multirow{2}{*}{\textsf{MRPC}} & Accuracy  & 0.8884 $\pm$ 0.0011 & \textbf{0.8915 $\pm$ 0.0026} & \textbf{0.0311} \\
                      & F1        & 0.8367 $\pm$ 0.0021 & \textbf{0.8431 $\pm$ 0.0073} & \textbf{0.022}  \\
\multirow{2}{*}{\textsf{QQP}}  & Accuracy  & 0.9028 $\pm$ 0.0008 & \textbf{0.9033 $\pm$ 0.001}  & 0.057           \\
                      & F1        & 0.87 $\pm$ 0.0013   & \textbf{0.8705 $\pm$ 0.0016} & \textbf{0.05}   \\
\textsf{RTE}                   & Accuracy  & 0.5870 $\pm$ 0.0126 & \textbf{0.6238 $\pm$ 0.0131} & \textbf{0.0002}
\end{tabular}%
}
\end{table}

\section{Discussion}
\label{sect:discussion}
In this paper, we delve into the integration of cognitive science principles into artificial intelligence (AI), specifically through the lens of Attention Schema Theory. We present Attention Schema-based Attention Controller, for more efficient control of attention particularly in transformers. ASAC is built on the premise that human attention is an active process involving internal models that manage cognitive resources. According to the AST, individuals create simplified representations of their attention systems, enabling them to prioritize and allocate cognitive resources effectively. This concept has been translated into ASAC, which employs a Vector Quantized Variational AutoEncoder as both an attention abstractor and controller within transformer architectures.

Our experimental results across various datasets in vision and NLP domains highlight ASAC’s effectiveness in improving classification accuracy and accelerating learning. For instance, when embedded within vision transformers, ASAC consistently outperformed baseline models across multiple datasets. These findings demonstrate that ASAC not only boosts classification accuracy but also expedites the learning process, showcasing its potential to enhance overall AI model performance.

One standout feature of ASAC is its robustness and generalization capabilities, particularly when exposed to noisy and out-of-distribution datasets. ASAC’s ability to maintain performance under such challenging conditions underscores its adaptability and resilience—traits essential for real-world applications where data can often be unpredictable and varied.

Moreover, ASAC has shown promising results in multi-task settings, effectively managing attention across different tasks. This capability aligns with the human brain’s cognitive flexibility, allowing seamless transitions between tasks and efficient resource allocation. Experimental results indicate that ASAC can dynamically adjust its attention allocation based on each task’s requirements, a significant advancement over traditional attention mechanisms that often operate with fixed patterns.

Overall, with some preliminary experiments, we also demonstrate that the ASAC architecture is versatile and adaptable in many scenarios with some mixed results here and there. ASAC shows enhanced resilience to adversarial attacks. ASAC also excels in scenarios involving efficient learning and knowledge transfer. The model demonstrates improved performance in few-shot learning tasks, effectively learning from limited examples.

The integration of ASAC into language models, specifically the \textsf{DistilBERT} model, further demonstrates its potential for enhancing NLP tasks. Despite the challenges associated with incorporating the VQVAE module into pre-trained models, the results indicate that ASAC can improve performance on benchmark datasets like GLUE. This integration highlights ASAC’s potential to enhance not only vision models but also language models, thereby broadening its applicability across different AI domains.

In future, we will address the current limitations of this study by effectively incorporating the attention controller into larger and pre-trained models like large language models (LLMs). Additionally, we aim to explore and develop better architectures that more closely mimic human cognition. These efforts will not only enhance the performance and applicability of ASAC but also push the boundaries of how cognitive science principles can be integrated into advanced AI systems, ensuring they perform robustly and efficiently across various domains and tasks.

In conclusion, the development of ASAC represents a significant step forward in integrating cognitive science principles into AI. The implications of ASAC extend beyond mere performance improvements; they also shed light on the potential for creating AI systems that more closely mimic human cognitive processes. As AI continues to evolve, approaches like ASAC will be instrumental in developing models that can adapt, learn efficiently, and perform robustly in diverse and unpredictable environments.

\bibliography{main}
\bibliographystyle{tmlr}

\clearpage       
\appendix

\section*{APPENDIX}

\section{Pseudo Code and Details of Design of Attention Controller}
We present the pseudo codes for the implementation of ASAC in Algorithm \ref{alg:asac_in_mha}. The VQVAE function
mentioned in Algorithm \autoref{alg:asac_in_mha} is further described in Algorithm \ref{alg:vqvae}.

\begin{algorithm}
\caption{ASAC plugged into multi-head attention}
\label{alg:asac_in_mha}
\begin{algorithmic}[1]

\STATEx \textbf{Input:} scaled dot product: $Z = \frac{QK^{T}}{\sqrt{d_k}}$, where $Q$ are Queries and $K$ are Keys in Multi-Head Attention
\STATEx \textbf{Output:} Reconstructed scaled dot product: $\hat{Z}$
\STATE $\hat{Z} \leftarrow VQVAE(Z)$
\STATE \textbf{return} Reconstructed scaled dot product $\hat{Z}$

\end{algorithmic}
\end{algorithm}

\begin{algorithm}
\caption{VQVAE as an Attention Controller}
\label{alg:vqvae}
\begin{algorithmic}[1]

\STATEx \textbf{Input:} scaled dot product: $Z$
\STATEx \textbf{Output:} Reconstructed scaled dot product: $\hat{Z}$

\STATE $Z_e \leftarrow \text{Encoder}(Z)$
\STATE $Z_e = \text{EncoderLayers}(Z)$
\STATE // EncoderLayers contain two linear layers separated by ReLU activation
\STATE $Z_v \leftarrow \text{VQ\_update}(Z_e)$
\STATE Compute distances: $d_{ij} = \| Z_e[:,i] - Z_v^{\text{prev}}[:,j] \|_2$
\STATE Find nearest indices: $j_i = \arg \min_j d_{ij}$
\STATE Update codebook: $Z_{v,\text{new}} = \beta Z_v^{\text{prev}} + (1-\beta) Z_e[:,j_i]$
\STATE \textbf{return} $Z_{v,\text{new}}$
\STATE $Z_d \leftarrow \text{Decoder}(Z_v)$
\STATE $Z_d = \text{DecoderLayers}(Z_v)$
\STATE // DecoderLayers contain two linear layers separated by ReLU activation
\STATE \textbf{return} Reconstructed scaled dot product $\hat{Z}$
\end{algorithmic}
\end{algorithm}

\section{Datasets}
\label{appx_datasets}
In this section, we describe all the datasets used in the paper. A summary of the vision datasets is available in \autoref{tab:dataset_stats}.

\begin{table}[h]
\centering
\caption{Statistics of Dataset}
\label{tab:dataset_stats}
\resizebox{\textwidth}{!}{%
\begin{tabular}{l|llllll}
Dataset       & Channels & Image size & Train samples & Test samples & Classes & Samples per class \\ \hline
\textsf{FashionMNIST}  & 1        & 28x28      & 60000         & 10000        & 10      & 7000              \\
\textsf{CIFAR-10}   & 3        & 32x32      & 50000         & 10000        & 10      & 6000              \\
\textsf{CIFAR-100}     & 3        & 32x32      & 50000         & 10000        & 100     & 600               \\
\textsf{SVHN}          & 3        & 32x32      & 73257         & 26032        & 10      & variable          \\
\textsf{Triangles}     & 1        & 64x64      & 5000000       & 1000000      & 2       & 2500000           \\
\textsf{Polygons}      & 1        & 64x64      & 5000000       & 1000000      & 2       & 2500000           \\
\textsf{Tiny Imagenet} & 3        & 64x64      & 100000        & 10000        & 200     & 500               \\
\textsf{Places365}     & 3        & 256x256    & 1803460       & 328500       & 365     & variable          \\
\textsf{CelebA}        & 3        & 218x178    & 162770        & 39829        & 40      & variable          \\
\textsf{Sort-of-Clevr} & 3        & 75x75      & 50000         & 10000        & 10      & 5000              \\
\textsf{MNIST}         & 1        & 28x28      & 60000         & 10000        & 10      & 6000              \\
\textsf{ODIR-5K}       & 3        & variable   & 2800          & 700          & 8       & variable          \\
\textsf{CIFAR10}-\textsf{SVHN}  & 3        & 32x32      & 123257        & 36032        & 10      & variable         
\end{tabular}%
}
\end{table}

\textbf{\textsf{Triangles}}: This dataset contains images of three white dot clusters forming a \textsf{Triangle} on a black background, as shown in \autoref{fig:Triangles_Polygons}(a). The task is to predict whether the Triangle is equilateral, making it a binary classification task. The images are 64 x 64 pixels, and the model must determine if the centroids of the white clusters are equidistant. The training data comprises 5 million pictures, and the test data contains 2 million images. We divide the image into 4 x 4 patches, each serving as a position in the input sequence for the architecture. The model must control its attention to ascertain the distance between the centroids.

\begin{figure}[h] 
    \centering
    \includegraphics[width=0.6\textwidth]{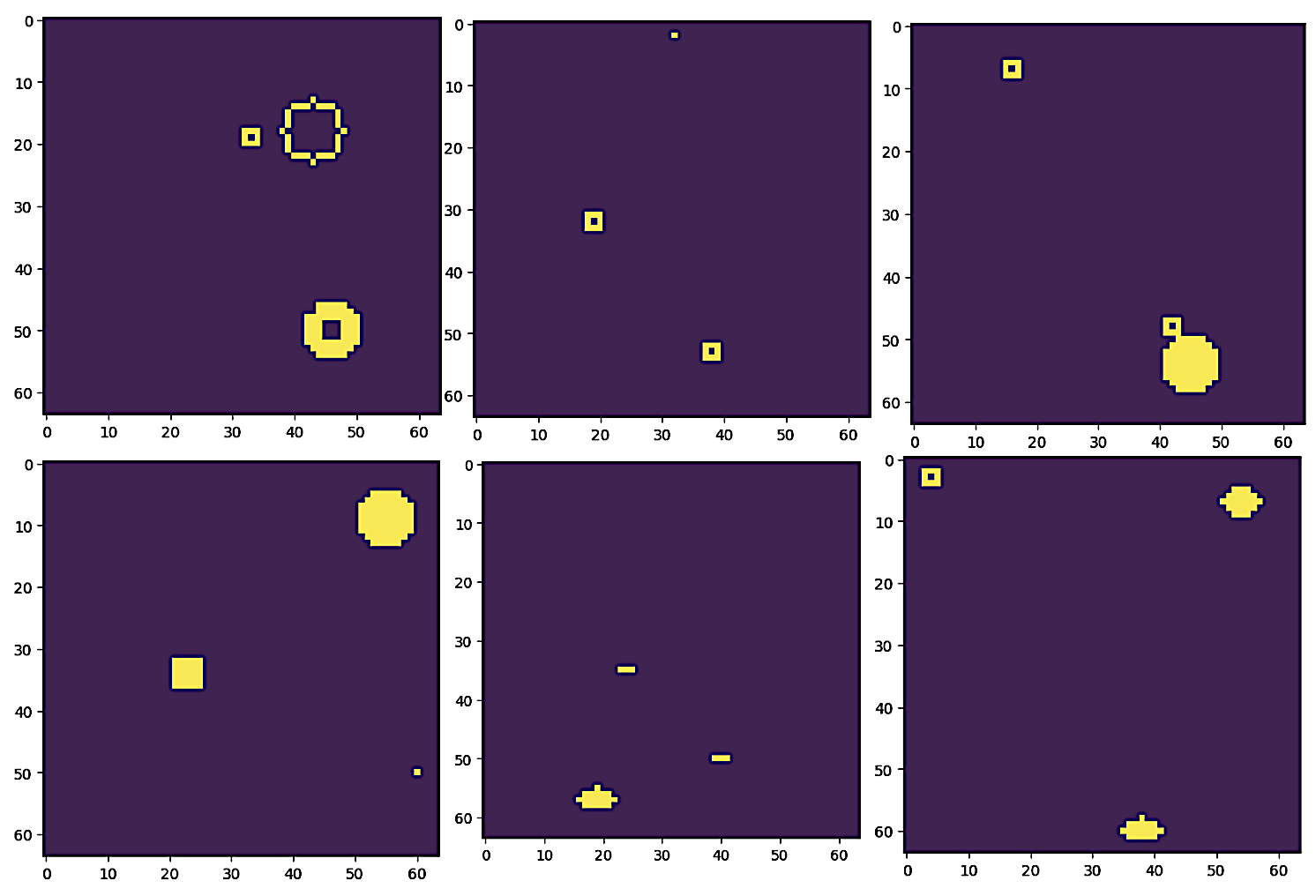} 
    \caption{Samples from \textsf{Triangles-OOD} data.}
    \label{Triangles_ood}
\end{figure}

For multi-task scenario, we use the same images but create different labels for different tasks. We create two tasks with \textsf{Triangles} dataset: one, to detect whether the clusters form an equilateral Triangle, and second, whether majority of the clusters are in the upper half of the image. Samples from the data are shown in \autoref{fig:multitask}. For Out-of-Distribution experiments, we generate images with different shapes, and sizes and fill content for the \textsf{Triangle} vertices. As shown in \autoref{Triangles_ood}, the shape can be a circle, Triangle, or square. The sizes can vary and the shape can be filled or empty.

\begin{figure}[h] 
    \centering
    \includegraphics[width=0.4\textwidth]{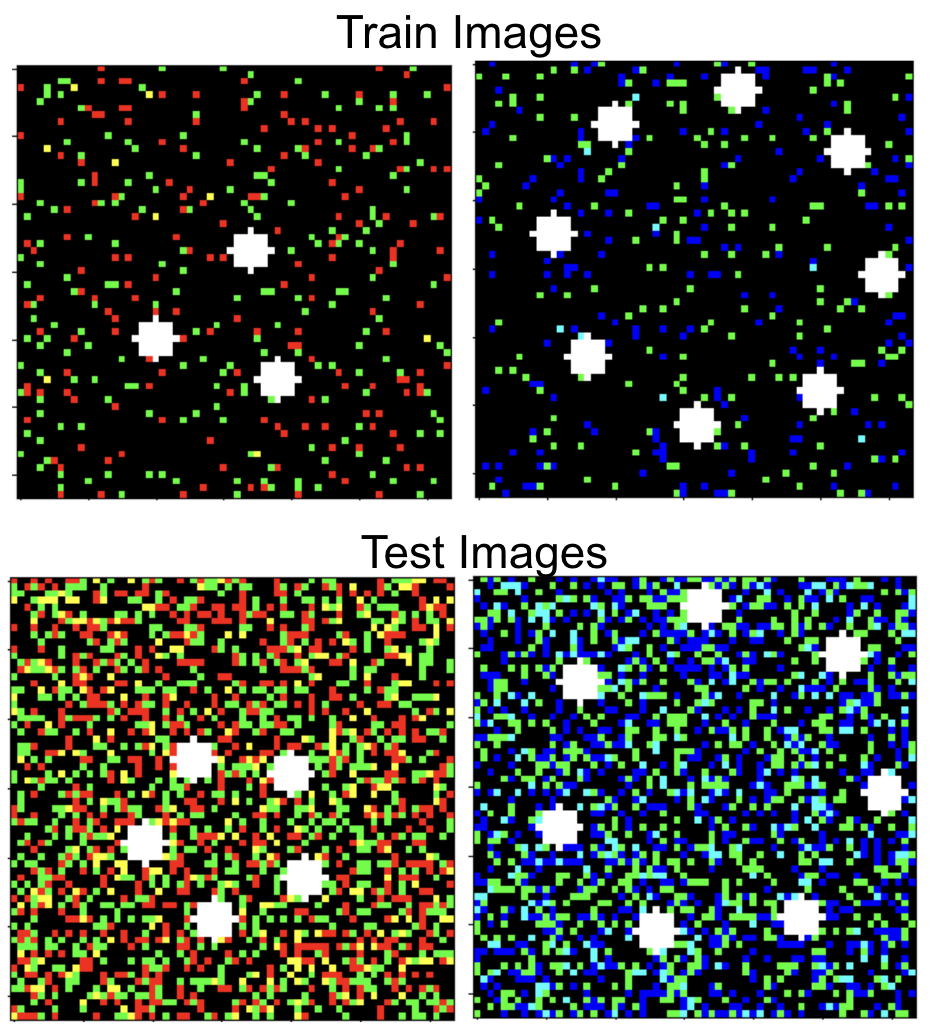} 
    \caption{Samples from \textsf{Polygons-OOD} data.}
    \label{fig:Polygons_ood}
\end{figure}

\textbf{\textsf{Polygons}}: The Regular \textsf{Polygons} dataset comprises images of Polygons created from dot clusters to predict whether a Polygon is regular, meaning all its sides are of equal length. The task is akin to identifying Equilateral \textsf{Triangles} but with added complexities such as background noise and alternating background and foreground colors. The images, which are 64 x 64 in size, can have between three and fifteen vertices. We show samples of images from the data in \autoref{fig:Triangles_Polygons}(b). The task involves binary image classification, with the images divided into equal-sized patches for the model.

For \textsf{Polygons-OOD} experiments, we slightly modify the train and test images as shown in \autoref{fig:Polygons_ood}. The train set contains \textsf{Polygons} with 3, 4, and vertices with colored noise covering 5\% of the background. The test set contains images of \textsf{Polygons} with 5, 6, and 7 vertices and the colored noise spans 25\% of the background.

\begin{figure}[h] 
    \centering
    \includegraphics[width=0.7\textwidth]{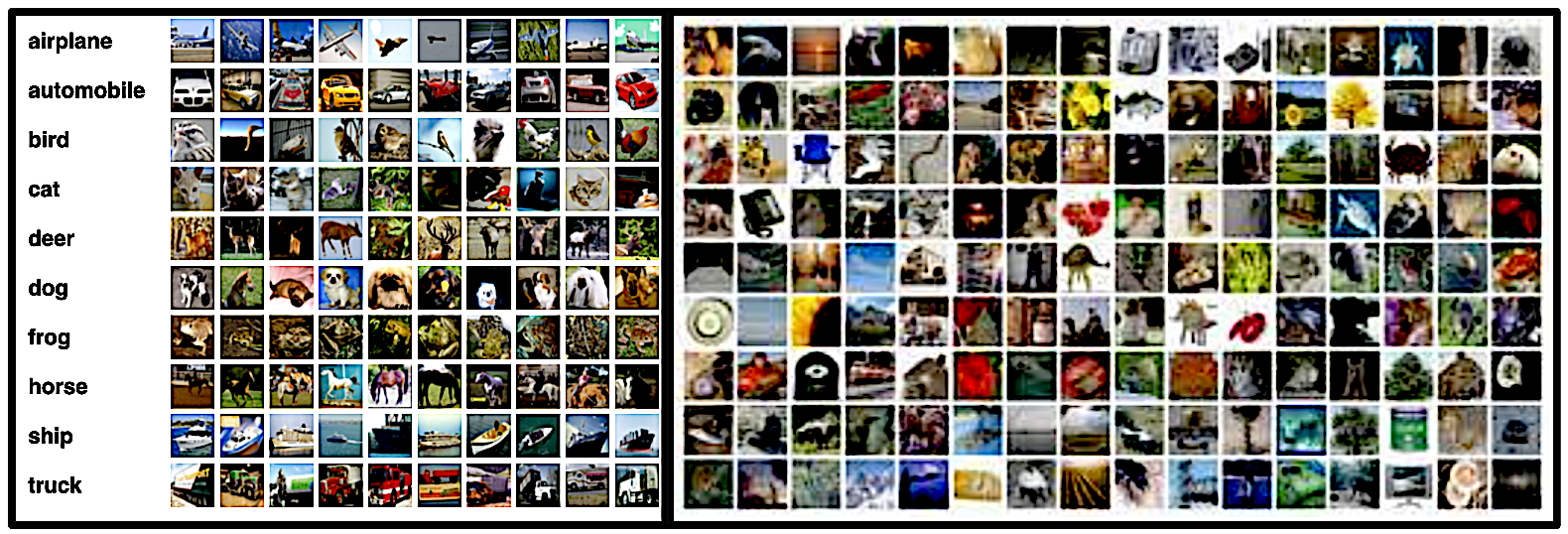} 
    \caption{Samples from \textsf{CIFAR10} (left) and \textsf{CIFAR-100} (right).}
    \label{fig:CIFAR10_100}
\end{figure}

\begin{figure}[h] 
    \centering
    \includegraphics[width=0.7\textwidth]{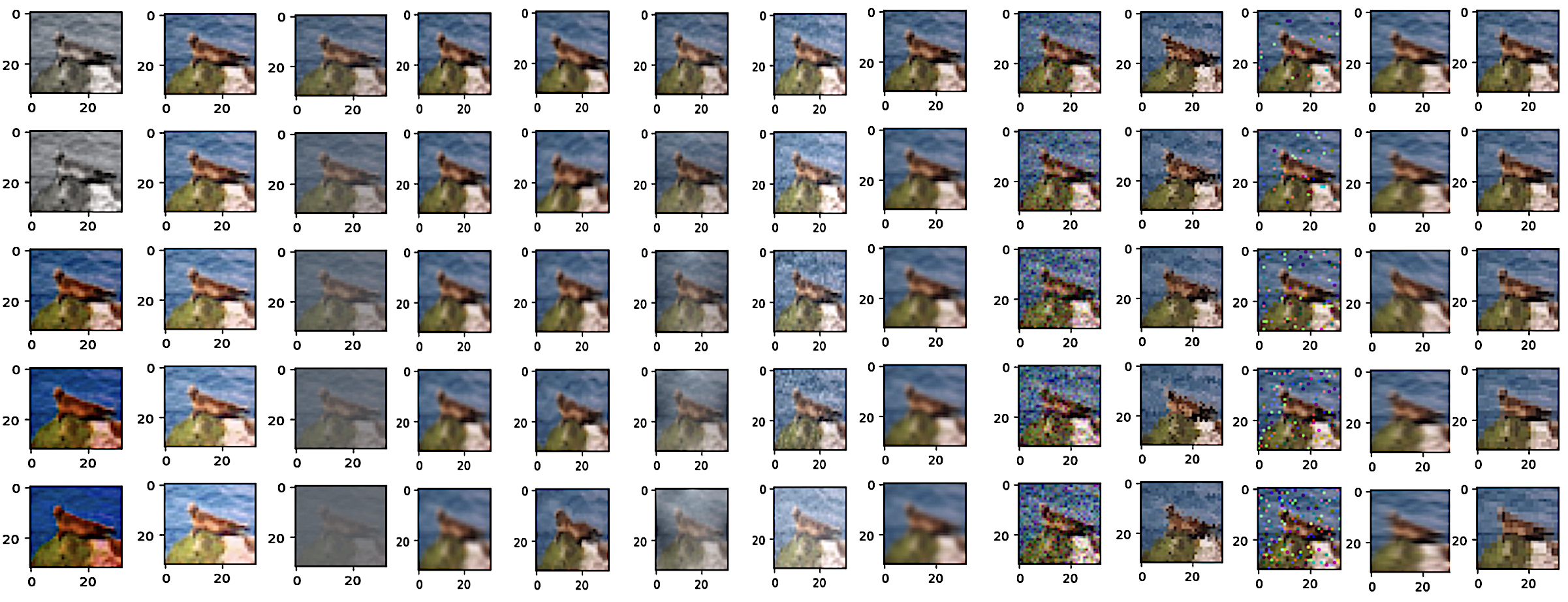} 
    \caption{Samples from Corrupted CIFAR data.}
    \label{fig:corrupted_cifar}
\end{figure}

\textbf{\textsf{CIFAR10}/\textsf{CIFAR100}}: \textsf{CIFAR-10} and \textsf{CIFAR-100} are labeled subsets of the 80 million tiny image dataset, featuring 32 x 32 color images of various objects for classification. \textsf{CIFAR-10} contains 60,000 images across ten classes, with 6,000 images per class. It includes 50,000 training images and 10,000 test images. \textsf{CIFAR-100}, similar to \textsf{CIFAR-10}, has 100 categories with 600 pictures each, where training and testing contain 500 and 100 images per class, respectively. Samples for both datasets are shown in \autoref{fig:CIFAR10_100}. These datasets are more challenging than the Equilateral \textsf{Triangles} or Regular \textsf{Polygons} tasks due to their color images and non-sparse objects. We divide the image data into equal-sized patches for input into the model.

For generalization experiments, we use corrupted version of \textsf{CIFAR10} dataset. This dataset has five levels of severity. Level 1 severity means lesser corruption, while level 5 has the most deterioration. The model is trained on the original \textsf{CIFAR-10} dataset and tested on different corruption levels of the \textsf{CIFAR-10-C}. Samples from dataset are shown in \autoref{fig:corrupted_cifar}.

\begin{figure}[h] 
    \centering
    \includegraphics[width=0.4\textwidth]{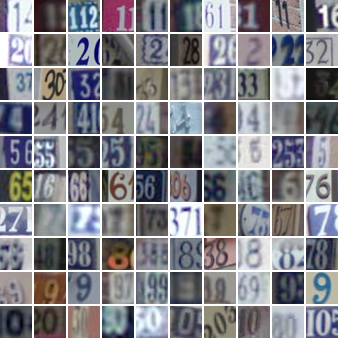} 
    \caption{Samples from \textsf{SVHN} data.}
    \label{fig:SVHN}
\end{figure}

\textbf{\textsf{SVHN}}:The Street View House Number (\textsf{SVHN}) is a real-world dataset obtained from house numbers in Google Street View images. We use the \textsf{MNIST}-like 32-by-32 images centered around a single character (many of the images do contain some distractors at the sides). This datasets has 10 classes, one for each digit. There are 73257 images and 26032 images in train and test set, respectively. The data is almost balanced, but it has a low degree of imbalance in the images. The samples are shown in \autoref{fig:SVHN}.

\textbf{\textsf{CIFAR10}-\textsf{SVHN}}: We mix the \textsf{CIFAR10} and \textsf{SVHN} dataset for multi-task experiments and analysis of usage of codebook.

\begin{figure}[h] 
    \centering
    \includegraphics[width=0.4\textwidth]{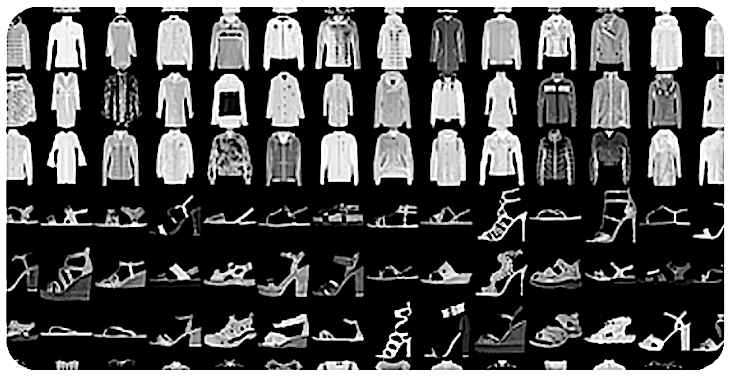} 
    \caption{Samples from \textsf{FashionMNIST} data.}
    \label{fig:FashionMNIST}
\end{figure}

\textbf{\textsf{FashionMNIST}}: The \textsf{FashionMNIST} dataset, featuring 28 x 28 grayscale images of 10 fashion items, is used for a multi-class classification task as shown in \autoref{fig:FashionMNIST}. It comprises a training set of 60,000 images and a test set of 10,000 images.

\begin{figure}[h] 
    \centering
    \includegraphics[width=0.5\textwidth]{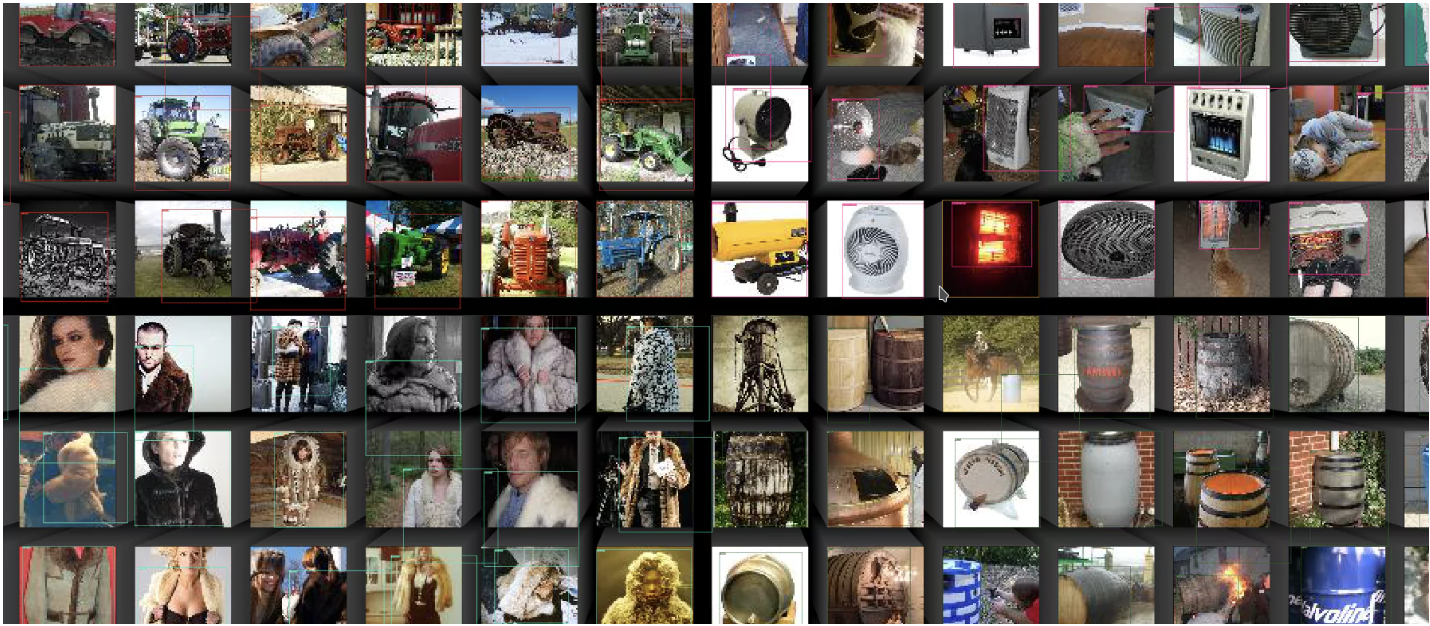} 
    \caption{Samples from \textsf{Tiny Imagenet} data.}
    \label{fig:tinyimagenet}
\end{figure}

\textbf{\textsf{Tiny Imagenet}}: \textsf{Tiny Imagenet} is a smaller version of Imagenet dataset. It contains 100,000 color images of 200 classes downsized to 64 x 64 pixels. Each class contains 500 training, 50 validation and 50 test images. Samples are shown in \autoref{fig:tinyimagenet}. We also use the corrupted \textsf{Tiny Imagenet} dataset for generalization experiments. The type of corruption and levels of corruption remain the same as that of \textsf{CIFAR10}-C.

\begin{figure}[h] 
    \centering
    \includegraphics[width=0.6\textwidth]{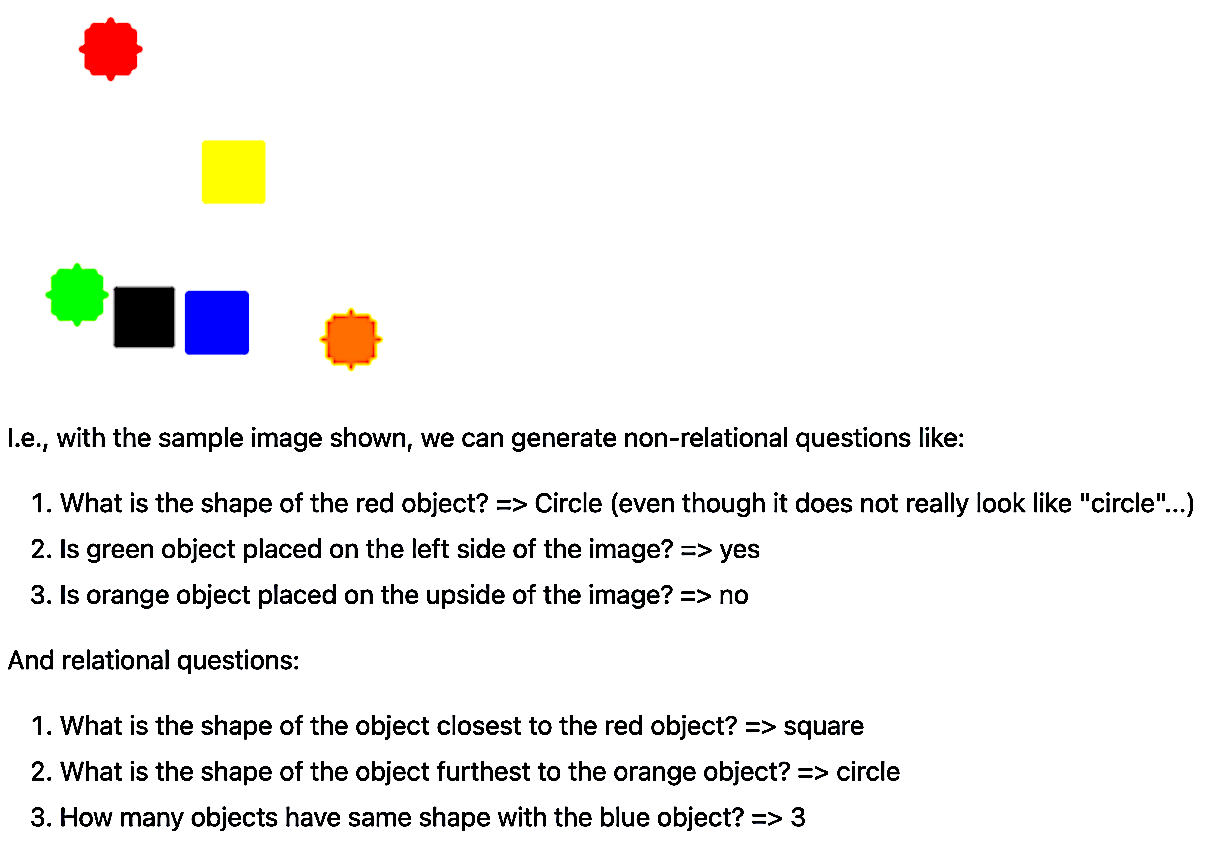} 
    \caption{Samples from \textsf{Sort-of-Clevr} data.}
    \label{fig:Sort-of-Clevr}
\end{figure}

\textbf{\textsf{Sort-of-Clevr}}: \textsf{Sort-of-Clevr} is a dataset designed to test reasoning capabilities inspired by the \textsf{CLEVR} dataset. It features images containing 2D objects of various shapes and colors and questions about the objects’ properties and their relationships to each other. The dataset has two tasks: relational and non-relational reasoning. Non-relational reasoning focuses on questions about a single object’s properties. The properties include the object’s shape and its horizontal and vertical location in the image. These questions do not require the model to consider relationships between multiple things. Alternatively, relational reasoning involves questions about the relationships between objects. These include identifying the object closest or furthest from another item or counting the number of objects sharing the same shape as another object. These questions necessitate the agent to consider the relations between different objects. Therefore, the \textsf{Sort-of-Clevr} dataset provides a comprehensive tool for evaluating an AI agent’s ability to reason about both individual objects and their relationships within a given context. A sample from the dataset is shown in \autoref{fig:Sort-of-Clevr}. We use the datatset in the multi-task experiments. In that experiment, the two tasks are relational and non-relational experiments within the \textsf{Sort-of-Clevr} dataset.

\begin{figure}[h] 
    \centering
    \includegraphics[width=0.5\textwidth]{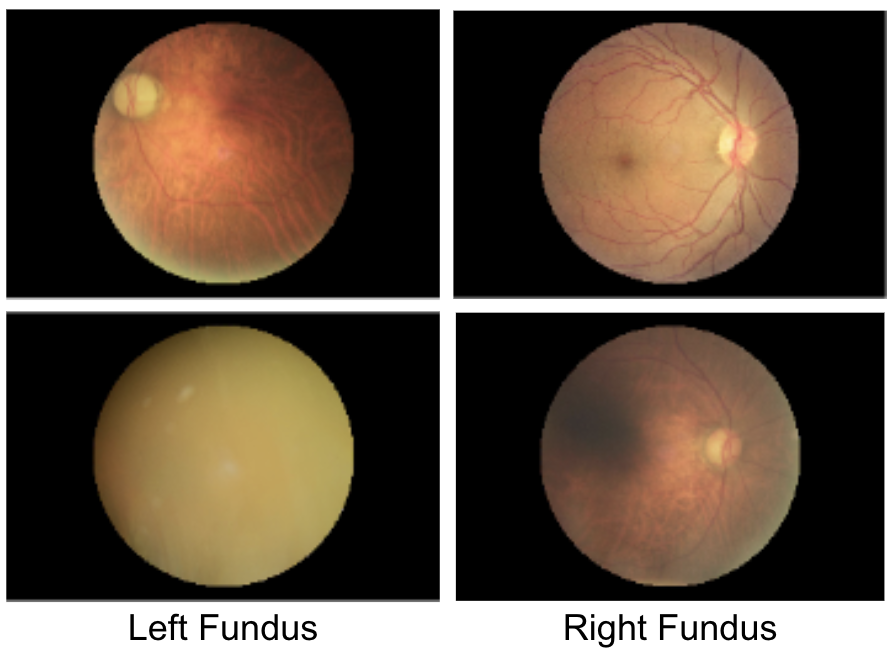} 
    \caption{Samples from \textsf{ODIR-5K} data.}
    \label{fig:ODIR-5K}
\end{figure}

\textbf{\textsf{Ocular Disease Intelligent Recognition}}: The \textsf{Ocular Disease Intelligent Recognition} (\textsf{ODIR-5K}) is a meticulously organized ophthalmic database comprising 5,000 patients, featuring age information, color fundus photographs of both left and right eyes, and diagnostic keywords provided by medical professionals. This database aims to encapsulate a practical collection of patient data sourced from a multitude of healthcare facilities in China, under the auspices of Shanggong Medical Technology Co., Ltd. Within these healthcare settings, fundus images are captured using a variety of commercially available cameras such as Canon, Zeiss, and Kowa, leading to variations in image resolutions. The annotations present in the dataset have been meticulously assigned by proficient human readers who have undergone rigorous quality control measures. These annotations serve the purpose of categorizing patients into eight distinct labels: Normal (N), Diabetes (D), Glaucoma (G), Cataract (C), Age-related Macular Degeneration (A), Hypertension (H), Pathological Myopia (M), Other diseases/abnormalities (O). Samples of the images from the dataset are shown in \autoref{fig:ODIR-5K}.

\begin{figure}[h] 
    \centering
    \includegraphics[width=0.5\textwidth]{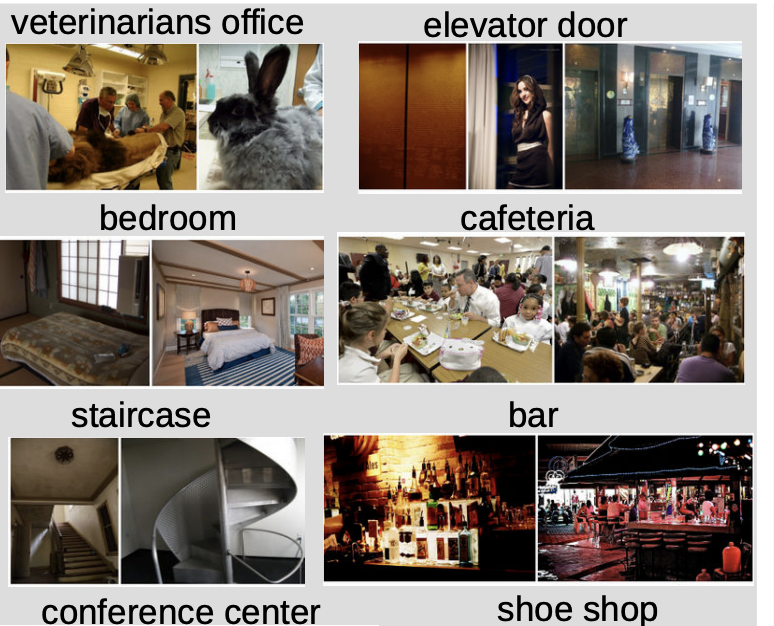} 
    \caption{Samples from \textsf{Places365} data.}
    \label{fig:Places365}
\end{figure}

\textbf{\textsf{Places365}}: We use \textsf{Places365-Standard} or \textsf{Places365-small} dataset, which is a scene recognition dataset. It contains 1.8 million train images from 365 scene categories. The train dataset contains 1803460 samples, and the test set contains 328500 samples. The dataset features 5000 to 30,000 training images per class, consistent with real-world frequencies of occurrence. We show the samples from the dataset in \autoref{fig:Places365}. We also use the same images for multi-task experiments with the following three tasks: first, classification into original labels, second, classification into level 1 hierarchy and third is the classification into level 2 hierarchy. The first task is single-label classification as opposed to the second and the third task which are multi-label classification. The level 1 hierarchy contains three labels: indoor, outdoor (natural), and outdoor (man-made). The level 2 hierarchy has 16 labels like shopping and dining, workplace, cultural, water, forest, transportation, sport fields, commercial buildings, etc.

\begin{figure}[h] 
    \centering
    \includegraphics[width=0.5\textwidth]{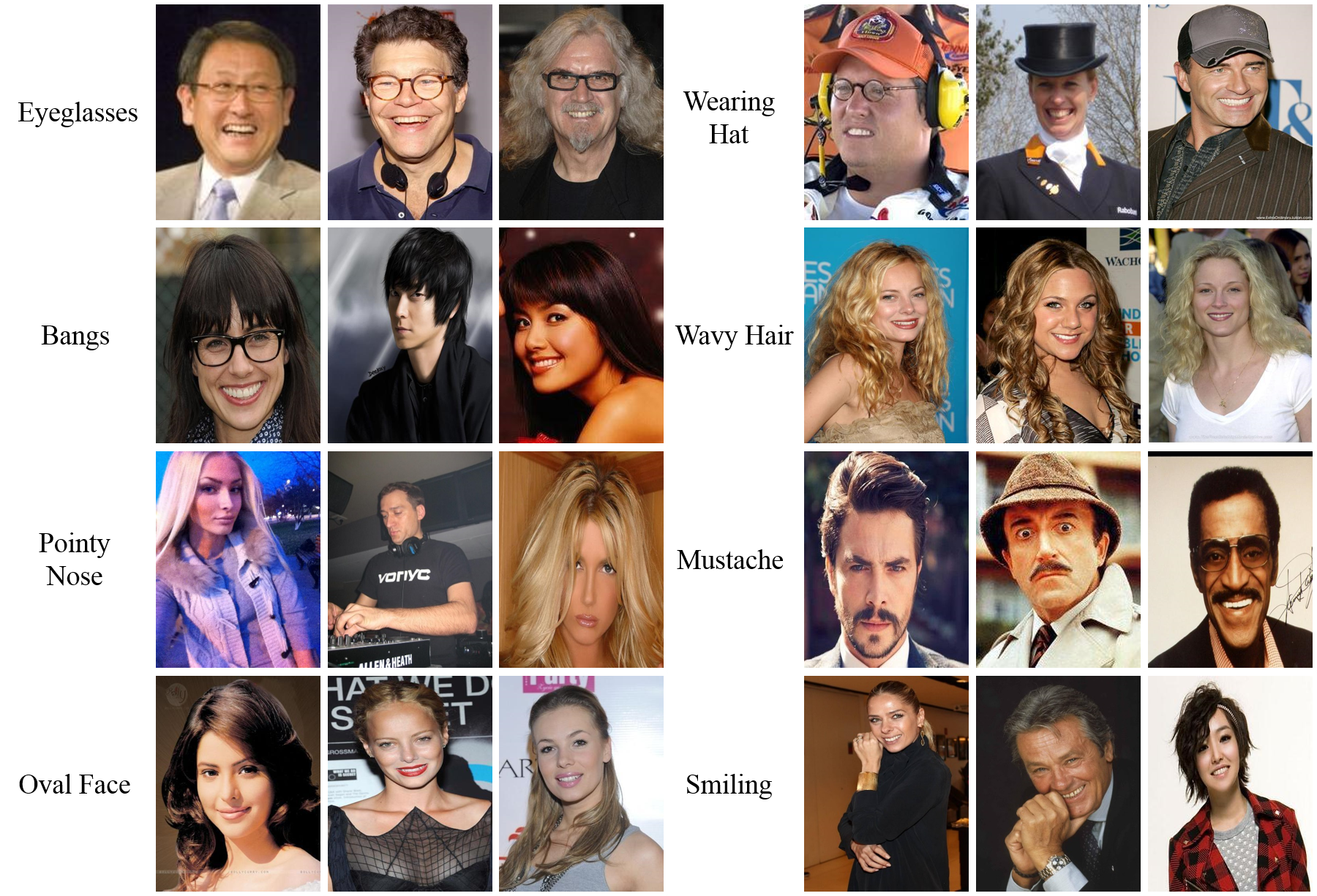} 
    \caption{Samples from \textsf{CelebA} data.}
    \label{fig:CelebA}
\end{figure}

\textbf{\textsf{CelebA}}: This is large-scale CelebFaces Attributes dataset. It has 202,599 number of face images, and 40 binary attributes annotations per image. The images in this dataset cover large pose variations and background clutter. This is a multi-label classification and samples from the datasets are shown in \autoref{fig:CelebA}.

\begin{figure}[h] 
    \centering
    \includegraphics[width=0.5\textwidth]{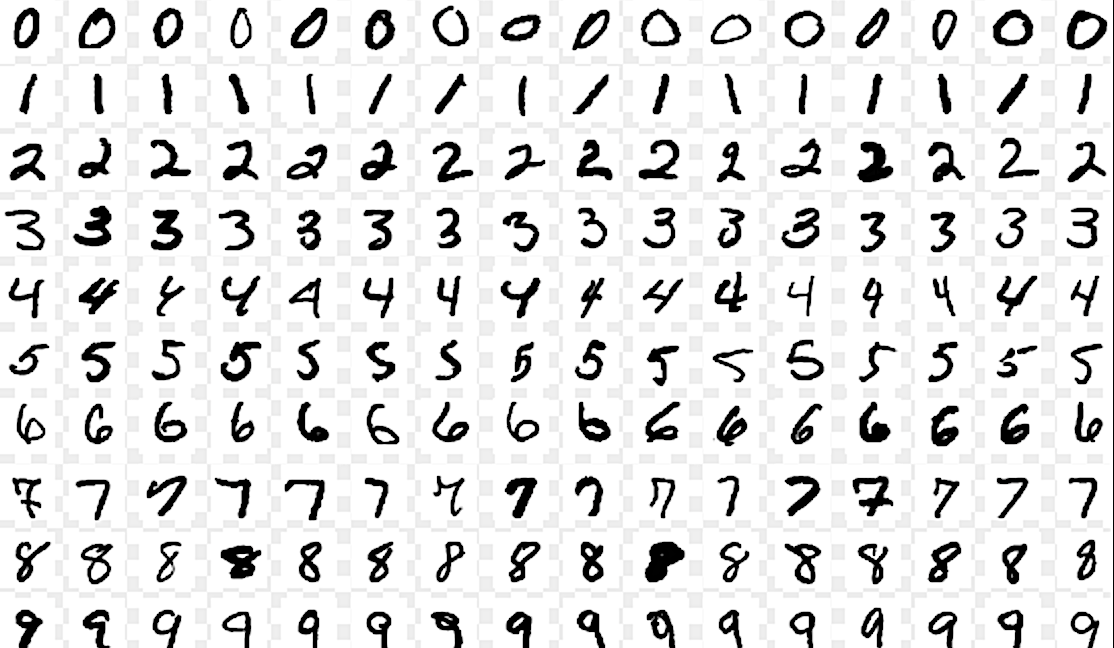} 
    \caption{Samples from \textsf{MNIST} data.}
    \label{fig:MNIST}
\end{figure}

\textbf{\textsf{MNIST}}: The \textsf{MNIST} database is a large database of handwritten digits. Samples from the dataset are shown in \autoref{fig:MNIST}. For the two-task scenario, we alter the labels of the data based on the tasks, the first one is to detect odd and even numbers and the second one is to detect prime and composite numbers. Just as we demonstrate in \textsf{Triangles} multi-task scenario above, here too, the task ID is 0 and 1 and the labels for the first task are 0 corresponding to even numbers and 1 for odd numbers. For task ID 1, the labels 0 and 1 correspond to prime and composite numbers, respectively.

\textbf{GLUE}: GLUE, the General Language Understanding Evaluation benchmark\footnote{\url{https://gluebenchmark.com/}} is a collection of resources for training, evaluating, and analyzing natural language understanding systems.  Specifically, we use the following datasets in the benchmark:
\begin{itemize}
    \item \textsf{STSB}: The Semantic Textual Similarity Benchmark is a collection of sentence pairs drawn from news headlines, video and image captions, and natural language inference data. Each pair is human-annotated with a similarity score from 1 to 5.
    \item \textsf{SST2}: The Stanford Sentiment Treebank consists of sentences from movie reviews and human annotations of their sentiment. The task is to predict the sentiment of a given sentence. It uses the two-way (positive/negative) class split, with only sentence-level labels.
    \item \textsf{QNLI}: The Stanford Question Answering Dataset is a question-answering dataset consisting of question-paragraph pairs, where one of the sentences in the paragraph (drawn from Wikipedia) contains the answer to the corresponding question (written by an annotator). The authors of the benchmark convert the task into sentence pair classification by forming a pair between each question and each sentence in the corresponding context, and filtering out pairs with low lexical overlap between the question and the context sentence. The task is to determine whether the context sentence contains the answer to the question. This modified version of the original task removes the requirement that the model select the exact answer, but also removes the simplifying assumptions that the answer is always present in the input and that lexical overlap is a reliable cue.
    \item \textsf{MRPC}: The Microsoft Research Paraphrase Corpus is a corpus of sentence pairs automatically extracted from online news sources, with human annotations for whether the sentences in the pair are semantically equivalent.
    \item \textsf{QQP}: The Quora Question Pairs2 dataset is a collection of question pairs from the community question-answering website Quora. The task is to determine whether a pair of questions are semantically equivalent.
    \item \textsf{RTE}: The Recognizing Textual Entailment (\textsf{RTE}) datasets come from a series of annual textual entailment challenges. Examples are constructed based on news and Wikipedia text. The authors of the benchmark convert all datasets to a two-class split, where for three-class datasets they collapse neutral and contradiction into not entailment, for consistency.
\end{itemize}

We show summary of GLUE benchmark in \autoref{tab:glue_dataset}.

\begin{table}[]
\centering
\caption{GLUE dataset summary}
\label{tab:glue_dataset}
\resizebox{0.75\textwidth}{!}{%
\begin{tabular}{l|l|l|l|l|l}
Dataset & Metrics            & Train & Validation & Labels & Distribution \\ \hline
\textsf{STSB}    & Pearson, spearmanR & 5.75k & 1.5k       & 0 to 5 & imbalanced   \\
\textsf{SST2}    & Accuracy           & 67.3k & 872        & 0,1    & imbalanced   \\
\textsf{QNLI}    & Accuracy           & 105k  & 5.46k      & 0,1    & balanced     \\
\textsf{MRPC}    & Accuracy, F1       & 3.67k & 408        & 0,1    & imbalanced   \\
\textsf{QQP}     & Accuracy, F1       & 364k  & 40.4k      & 0,1    & imbalanced   \\
\textsf{RTE}     & Accuracy           & 2.49k & 277        & 0,1    & balanced    
\end{tabular}%
}
\end{table}

\section{Details of the Experiments}
\subsection{Attention Control on Multi-task scenario}

\begin{figure}[h] 
    \centering
    \includegraphics[width=0.5\textwidth]{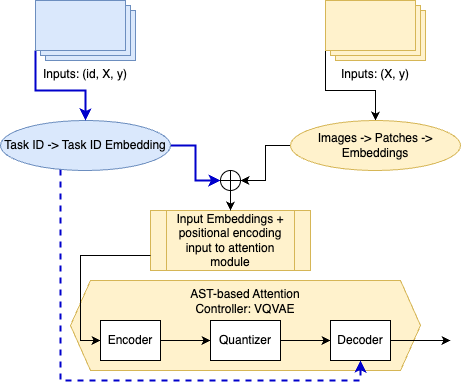} 
    \caption{Task ID integration in the input}
    \label{fig:taskid}
\end{figure}

In order to conduct the multi-task experiments, we include unique task identifiers in the dataset. These task identifiers (Task IDs) serve to distinguish between different types of tasks. Although the dataset remains constant, the corresponding labels may vary depending on the task identifiers. During the classification trials involving an attention controller, the original input images undergo a transformation into a sequence of distinct patches. Subsequently, these patches are converted into patch embeddings, and positional embeddings are computed, ultimately forming the final input sequence that is fed into the model. In the context of a multi-task setting, we augment the input sequence with a task embedding before passing it to the model for processing. As shown in \autoref{fig:taskid}, the (right) yellow part is the standard way of processing the input when the input only contains input data and labels (X, y). If the input data are images, then X are images which are converted to a sequence of patches. Patch embeddings and their positional encodings are calculated and fed to the transformers attention module. In the multi-task scenarios, the dataset contains an extra information of task ID: (id, X, y). The task ID is used to calculate the Task ID embeddings (left, blue part in the \autoref{fig:taskid}) which are concatenated to the input embeddings. \autoref{fig:taskid} also show the variations that we mention in the main content of how the task information is passed into the model.

\section{Hyperparameters for different experiments}
We ran all the experiments on NVIDIA V100 16GB GPU from Google Cloud Platform and all the codes were written in PyTorch. The following hyperparameters are common in all the experiments: $patch\_size= 4, \lambda = 0.01, train\_batch\_size = 64, eval\_batch\_size = 64, learning\_rate = 0.0001, weight\_decay = 0.01, hidden\_dropout\_prob = 0.1, attention\_probs\_dropout\_prob = 0.1, threshold\_ema\_dead\_code = 2, commitment\_cost = 1.0$. The other hyperparameters are shown in \autoref{tab:hyperparameters1}. For noisy and OOD experiments, the hyperparameters for \textsf{CIFAR-C}, \textsf{Tiny Imagenet}-C, \textsf{Triangles-OOD}, \textsf{Polygons-OOD} experiments remain the same as classification experiment hyperparameters of that of their non-corrupted datasets shown in the \autoref{tab:hyperparameters1}. Hyperparameters for \textsf{DistilBERT} experiments are available in \autoref{tab:hyperparametersDistilBERT}. Other parameters apart from VQVAE-specific parameters remain the same in \textsf{DistilBERT} model. We performed grid search on several values of hyperparameters and provide the hyperparameter with best results presented in the main content.

We show the hyperparameters for multi-task experiments in \autoref{tab:hyperparameters_multitask}. The following hyperparameters are common in all the multi-task experiments: $patch\_size = 4, \lambda = 0.01, train\_batch\_size = 64, eval\_batch\_size = 64, learning\_rate = 0.0001, weight\_decay = 0.01, hidden\_dropout\_prob = 0.1, attention\_probs\_dropout\_prob = 0.1, threshold\_ema\_dead\_code = 2, commitment\_cost = 1.0$. In all the vision experiments, the total number of parameters in the baseline and ASAC remains almost equal.

\begin{table}[]
\centering
\caption{Hyperparameters for classification and generalization experiments. “cb" stands for codebook.}
\label{tab:hyperparameters1}
\resizebox{\textwidth}{!}{%
\begin{tabular}{l|l|lllllll}
Dataset       & Experiment & layer & head & h\_dim & ffn\_dim & latent\_dim & cb\_dim & cb\_size \\ \hline
\textsf{CIFAR10}       & Baseline   & 4     & 4    & 368    & 532      & -           & -       & -        \\
              & ASAC       & 4     & 4    & 256    & 512      & 400         & 64      & 256      \\
\textsf{CIFAR100}      & Baseline   & 4     & 8    & 632    & 1044     & -           & -       & -        \\
              & ASAC       & 4     & 8    & 512    & 1024     & 400         & 64      & 512      \\
\textsf{SVHN}          & Baseline   & 4     & 4    & 368    & 532      & -           & -       & -        \\
              & ASAC       & 4     & 4    & 256    & 512      & 400         & 64      & 256      \\
\textsf{FashionMNIST}  & Baseline   & 4     & 8    & 264    & 510      & -           & -       & -        \\
              & ASAC       & 4     & 8    & 128    & 256      & 320         & 32      & 128      \\
\textsf{Triangle}      & Baseline   & 4     & 4    & 528    & 544      & -           & -       & -        \\
              & ASAC       & 4     & 4    & 256    & 512      & 512         & 32      & 128      \\
\textsf{Polygon}       & Baseline   & 4     & 8    & 648    & 914      & -           & -       & -        \\
              & ASAC       & 4     & 8    & 256    & 512      & 512         & 32      & 128      \\
\textsf{Tiny Imagenet} & Baseline   & 4     & 4    & 524    & 548      & -           & -       & -        \\
              & ASAC       & 4     & 4    & 256    & 512      & 512         & 32      & 128      \\
\textsf{SOC (rel)}     & Baseline   & 4     & 4    & 524    & 564      & -           & -       & -        \\
              & ASAC       & 4     & 4    & 256    & 512      & 512         & 32      & 128      \\
\textsf{SOC (no rel)}  & Baseline   & 4     & 4    & 254    & 564      & -           & -       & -        \\
              & ASAC       & 4     & 4    & 256    & 512      & 512         & 32      & 128      \\
\textsf{Places365}     & Baseline   & 6     & 8    & 728    & 1384     & -           & -       & -        \\
              & ASAC       & 6     & 8    & 512    & 1024     & 400         & 256     & 1024     \\
\textsf{CelebA}        & Baseline   & 4     & 4    & 528    & 544      & -           & -       & -        \\
              & ASAC       & 4     & 4    & 256    & 512      & 512         & 32      & 128     
\end{tabular}%
}
\end{table}

\begin{table}[]
\centering
\caption{Hyper-parameters for \textsf{DistilBERT} experiments}
\label{tab:hyperparametersDistilBERT}
\resizebox{0.75\textwidth}{!}{%
\begin{tabular}{l|l|lll}
Dataset & Model    & codebook\_dim & latent\_dim & codebook\_size \\ \hline
\textsf{STSB}    & Baseline & 256           & 128         & 128            \\
        & ASAC     & 256           & 256         & 512            \\
\textsf{SST2}    & Baseline & 128           & 256         & 512            \\
        & ASAC     & 128           & 256         & 512            \\
\textsf{QNLI}    & Baseline & 128           & 512         & 128            \\
        & ASAC     & 128           & 512         & 128            \\
\textsf{MRPC}    & Baseline & 512           & 256         & 512            \\
        & ASAC     & 512           & 256         & 512            \\
\textsf{QQP}     & Baseline & 512           & 786         & 786            \\
        & ASAC     & 786           & 786         & 1024           \\
\textsf{RTE}     & Baseline & 256           & 128         & 128            \\
        & ASAC     & 256           & 128         & 128           
\end{tabular}%
}
\end{table}

\begin{table}[]
\centering
\caption{Hyperparameters for Multi-task experiments. “cb" stands for codebook.}
\label{tab:hyperparameters_multitask}
\resizebox{\textwidth}{!}{%
\begin{tabular}{l|l|lllllll}
Dataset    & Experiment        & layer & head & h\_dim & ffn\_dim & latent\_dim & cb\_dim & cb\_size \\ \hline
\textsf{MNIST}      & Baseline          & 4     & 4    & 248    & 272      & -           & -       & -        \\
           & TaskID in input   & 4     & 4    & 152    & 256      & 320         & 32      & 128      \\
           & TaskID in decoder & 4     & 4    & 128    & 260      & 320         & 32      & 128      \\
           & TaskID in both    & 4     & 4    & 128    & 250      & 320         & 32      & 128      \\
\textsf{ODIR-5K}    & Baseline          & 4     & 8    & 200    & 400      & -           & -       & -        \\
           & TaskID in input   & 4     & 8    & 64     & 128      & 64          & 32      & 64       \\
           & TaskID in decoder & 4     & 8    & 64     & 104      & 64          & 32      & 64       \\
           & TaskID in both    & 4     & 8    & 64     & 98       & 64          & 32      & 64       \\
\textsf{Triangle}   & Baseline          & 4     & 4    & 528    & 544      & -           & -       & -        \\
           & TaskID in input   & 4     & 4    & 256    & 506      & 512         & 32      & 128      \\
           & TaskID in decoder & 4     & 4    & 232    & 416      & 512         & 32      & 128      \\
           & TaskID in both    & 4     & 4    & 232    & 406      & 512         & 32      & 128      \\
\textsf{SOC}        & Baseline          & 4     & 4    & 528    & 544      & -           & -       & -        \\
           & TaskID in input   & 4     & 4    & 256    & 506      & 512         & 32      & 128      \\
           & TaskID in decoder & 4     & 4    & 232    & 416      & 512         & 32      & 128      \\
           & TaskID in both    & 4     & 4    & 232    & 406      & 512         & 32      & 128      \\
\textsf{CIFAR-SVHN} & Baseline          & 4     & 4    & 300    & 604      & -           & -       & -        \\
           & TaskID in input   & 4     & 4    & 320    & 610      & 128         & 64      & 1024     \\
           & TaskID in decoder & 4     & 4    & 312    & 596      & 128         & 64      & 1024     \\
           & TaskID in both    & 4     & 4    & 312    & 596      & 128         & 64      & 1024     \\
\textsf{Places365}  & Baseline          & 6     & 8    & 728    & 1384     & -           & -       & -        \\
           & TaskID in both    & 6     & 8    & 488    & 970      & 400         & 256     & 1024    
\end{tabular}%
}
\end{table}

\end{document}

%% file: math_commands.tex
\usepackage{amsmath,amsfonts,bm}

\def\eqref#1{equation~\ref{#1}}

\def\1{\bm{1}}

\DeclareMathAlphabet{\mathsfit}{\encodingdefault}{\sfdefault}{m}{sl}
\SetMathAlphabet{\mathsfit}{bold}{\encodingdefault}{\sfdefault}{bx}{n}

